\documentclass[11pt]{article}

\usepackage[preprint]{acl}

\usepackage{times}
\usepackage{latexsym}

\usepackage[T1]{fontenc}

\usepackage[utf8]{inputenc}

\usepackage{microtype}

\usepackage{inconsolata}

\usepackage{multirow, multicol}
\usepackage{array}      
\usepackage{booktabs}   
\usepackage{multirow}  
\usepackage{colortbl}   
\usepackage{xcolor}     
\usepackage{fontawesome5}
\usepackage{textgreek}
\usepackage{amsmath,amssymb,amsfonts}
\usepackage{graphicx}
\usepackage{array} 
\usepackage{float}    
\usepackage{enumitem}
\usepackage{lipsum}
\usepackage{multicol}
\usepackage{longtable}
\usepackage{multirow}
\usepackage{geometry}
\usepackage{tabularx}
\usepackage{makecell}
\usepackage{amsthm}
\usepackage{mathrsfs}
\usepackage[title]{appendix}
\usepackage{textcomp}
\usepackage{manyfoot}
\usepackage{hyperref}

\usepackage{algorithm}
\usepackage[noend]{algpseudocode} 

\algrenewcommand\algorithmicrequire{\textbf{Input:}}
\algrenewcommand\algorithmicensure{\textbf{Output:}}

\title{Syzygy of Thoughts: Improving LLM CoT with the Minimal Free Resolution}

\author{
  \textbf{Chenghao Li\textsuperscript{1}},
  \textbf{Chaoning Zhang\textsuperscript{1}\thanks{Corresponding author}},
  \textbf{Yi Lu\textsuperscript{2}},\\
  \textbf{Jiaquan Zhang\textsuperscript{1}},
  \textbf{Qigan Sun\textsuperscript{3}},
  \textbf{Xudong Wang\textsuperscript{3}},\\
  \textbf{Jiwei Wei\textsuperscript{1}},
  \textbf{Guoqing Wang\textsuperscript{1}},
  \textbf{Yang Yang\textsuperscript{1}},
  \textbf{Heng Tao Shen\textsuperscript{4,1}}
\\
\\
  \textsuperscript{1}UESTC;
  \textsuperscript{2}CNU;
  \textsuperscript{3}KHU;
  \textsuperscript{4}Tongji Univ.
}

\begin{document}
\maketitle

\begin{abstract}
Chain-of-Thought prompting enhances the reasoning of large language models by decomposing problems into sequential steps, mimicking human logic and reducing errors. However, for complex tasks with vast solution spaces and ambiguous constraints, a single reasoning chain often proves insufficient.
Inspired by Minimal Free Resolutions (MFR) in commutative algebra and algebraic geometry, we propose \textbf{Syzygy of Thoughts (SoT)}---a new framework that extends CoT by introducing auxiliary, interrelated reasoning paths. By explicitly modeling dependencies among these paths, SoT captures deeper logical structure and enables more robust, structured problem solving.
MFR decomposes a module into a sequence of free modules of minimal rank, providing a principled way to analyze complex systems. Drawing on MFR's principle of ``minimal representation, precise constraints, and layer-by-layer relation tracing,'' SoT systematically decomposes an original complex problem into logically complete minimal subproblems while preserving key problem characteristics and reducing reasoning length.
We evaluate SoT across diverse datasets (e.g., GSM8K, MATH) and model families (e.g., GPT-4o-mini, Qwen2.5). Results show that SoT achieves inference accuracy comparable to or exceeding mainstream CoT baselines. Moreover, by aligning the sampling process with algebraic constraints, SoT improves the scalability of inference-time computation, offering both transparent reasoning and strong performance. The code is available at~\href{https://github.com/dlMARiA/Syzygy-of-thoughts}{https://github.com/dlMARiA/Syzygy-of-thoughts}
\end{abstract}
\begin{figure}
    \centering
    \includegraphics[width=0.95\linewidth]{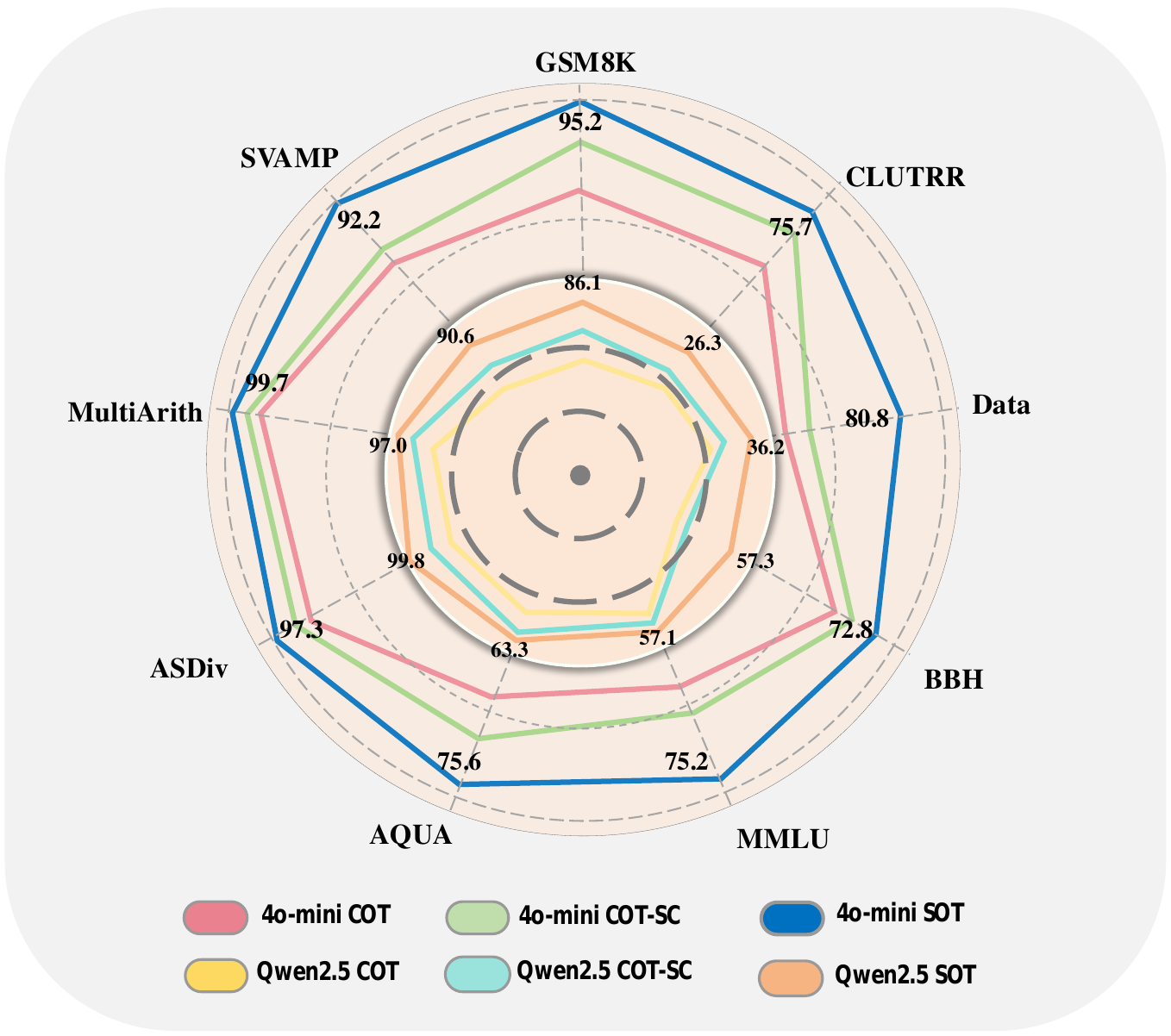}
    \caption{SoT (Ours) achieved performance improvements compared to CoT and CoT-SC on two models across nine datasets. The inner circle shows three methods of Qwen2.5, while the outer circle shows three methods of 4o-mini.}
    \label{fig:a}
\end{figure}

\section{Introduction}
\label{sec:intro}

The rapid advancements in natural language processing~\cite{vaswani2017attention,devlin2019bert,radford2018improving,brown2020language} have demonstrated outstanding reasoning capabilities across various tasks~\cite{zhao2023survey,min2023recent}. In recent years, the Chain of Thought (CoT) prompting method~\cite{wei2022chain} has emerged as an effective strategy for simulating human reasoning by generating intermediate reasoning steps, significantly enhancing model performance in solving complex problems. The core advantage of the CoT framework lies in its ability to decompose problems into a series of explicit and easily interpretable intermediate steps, thereby improving the transparency and explainability of the reasoning process. This explicit stepwise reasoning not only facilitates in-depth analysis and debugging of model outputs but also provides a clear logical pathway for solving complex problems.

However, despite its success in many tasks, CoT faces inherent limitations with high-dimensional, nonlinear, and abstract logical problems~\cite{wang2022towards,shi2023large,jin2024impact}. Traditional CoT methods rely on fixed or heuristic decompositions, which fail to capture essential details in multidimensional problems~\cite{wang2022towards}. Additionally, the lack of structured planning in intermediate steps leads to redundant computations and increased inference latency~\cite{jin2024impact}. While improvements like multi-path exploration, voting, and hierarchical strategies have helped, they still fail to fully address logical incompleteness and poor high-dimensional decomposition. Thus, challenges in robustness and accuracy remain for complex reasoning tasks.

To address these challenges, we propose a new reasoning framework—Syzygy of Thoughts (SoT). 
``Syzygy'' is derived from the Greek word \textit{syzygia}
($\sigma\upsilon\zeta\upsilon\gamma\iota\alpha$), meaning ``union'' or ``pairing''.
This method draws inspiration from the concept of MFR in algebraic geometry and computational algebra~\cite{eisenbud2013commutative,rossi2009minimal,botbol2021simplest}, systematically decomposing and reconstructing problems by constructing minimal free module sequences. The incorporation of MFR theory ensures a more rigorous and efficient problem decomposition process while preserving the core characteristics of the problem, effectively reducing redundant computations and logical inconsistencies. Specifically, SoT retains the stepwise reasoning advantages of CoT while introducing algebraic tools such as "module", "Betti numbers", "freeness", "mapping", "exactness" and "minimality" to deeply explore and systematically express the intrinsic structure of complex problems. This approach provides a new theoretical perspective for analyzing high-dimensional and multivariable problems.

To validate the effectiveness of SoT, we conducted extensive experiments across diverse datasets, including GSM8K, MATH, and others, spanning mathematical reasoning and high-dimensional problem domains. These datasets, with their complex structures and multifaceted features, provide a robust testbed for evaluating reasoning frameworks in real-world scenarios. Experimental results show that SoT achieves inference accuracy that matches or surpasses mainstream CoTs standards, demonstrating its superior performance and practical potential.

The main contributions of this work:
\begin{itemize}
\item We introduce SoT, an MFR-inspired inference-time reasoning framework that generalizes CoT from a single linear chain to interdependent, syzygy-structured reasoning paths.
\item We formalize the reasoning process through an MFR lens by mapping algebraic notions (Module, Betti numbers, Freeness, Mapping, Exactness, Minimality) to concrete prompting and search operations, enabling structured decomposition, logical-closure verification, and redundancy pruning.
\item We propose an algebraically constrained sampling/selection procedure that improves the accuracy--cost trade-off and provides stable inference under sampling randomness.
\item We conduct extensive experiments across nine benchmarks and multiple LLM backbones, showing that SoT consistently outperforms CoT, CoT-SC, GoT, and AoT, with lower or comparable variance, supported by cost and ablation analyses.
\end{itemize}

\section{Related Work}
\label{related}

\subsection{Advanced Reasoning Techniques in LLMs}
Recent advances in prompting have substantially improved the reasoning of LLMs, with Chain-of-Thought (CoT) prompting~\cite{wei2022chain} being a seminal example. CoT elicits step-by-step solutions but can still struggle on multidimensional or nonlinear tasks (e.g., abstract logic and advanced mathematics). To mitigate these issues, extensions such as Zero-shot-CoT~\cite{kojima2022large}, Self-Consistency CoT~\cite{wang2022self}, and Auto-CoT~\cite{zhang2022automatic} have been proposed, together with verification-based methods including VerifyCoT~\cite{zhao2023verify} and CoF-CoT~\cite{nguyen2023cof}.
Beyond linear CoT, structured prompting further strengthens reasoning. Least-to-Most Prompting (LtM)~\cite{zhou2022least} decomposes tasks into simpler subproblems, while programmatic paradigms---Program of Thought (PoT)~\cite{chen2022program}, Chain of Code (CoC)~\cite{li2023chain}, and Buffer of Thought (BoT)~\cite{yang2025buffer}---integrate discrete variables and procedural execution. Algorithm of Thought (AoT)~\cite{sel2023algorithm} then iteratively synthesizes and refines intermediate results to improve token efficiency.
Search-based paradigms explore multiple solution paths. Tree-of-Thought (ToT)~\cite{yao2023tree} performs hierarchical branching but may suffer from exponential growth, whereas Graph-of-Thought (GoT)~\cite{besta2024graph} generalizes this to flexible graph structures for aggregation and recalibration. Complementary approaches such as Skeleton-of-Thought (SoT)~\cite{ning2023skeleton}, Self-Refine~\cite{madaan2023self}, and Step-Back Prompting~\cite{zheng2023take} emphasize outlining and iterative refinement to improve accuracy and stability.

\subsection{Minimal Free Resolution}
MFR is a fundamental tool in homological algebra~\cite{fieldsteel2021minimal} and algebraic geometry~\cite{botbol2021simplest} for characterizing module structures~\cite{eisenbud2013commutative,rossi2009minimal} and deriving invariants such as rank, symmetry, and relations~\cite{chen2011hilbert}. In computational algebraic geometry, syzygy-based MFR supports the study of singularities and invariants of varieties~\cite{evans1981syzygy} and improves the complexity of Gr\"obner-basis computations~\cite{eisenbud2001computations,stewart1993early,weispfenning1992comprehensive,capani1997computing}.
Beyond pure algebra, MFR has been used to speed up persistent-homology computation in TDA~\cite{wasserman2018topological} and to analyze structures arising in physics and bioinformatics, including Calabi--Yau singularities~\cite{polchinski1994string}, gauge fields~\cite{hitchin2003generalized}, and gene regulatory networks~\cite{li2011stability}.
These strengths---exposing structured dependencies while reducing computational redundancy---motivate our integration of MFR into CoT for LLM reasoning. By decomposing symbolic dependencies among intermediate steps, we prune redundant branches, streamline intermediate computation, and improve the transparency and efficiency of reasoning, providing a more structured pathway for solving complex problems.
\section{Method}
\label{sec:method}

\begin{figure}[t!]
    \centering
    \includegraphics[width=\linewidth]{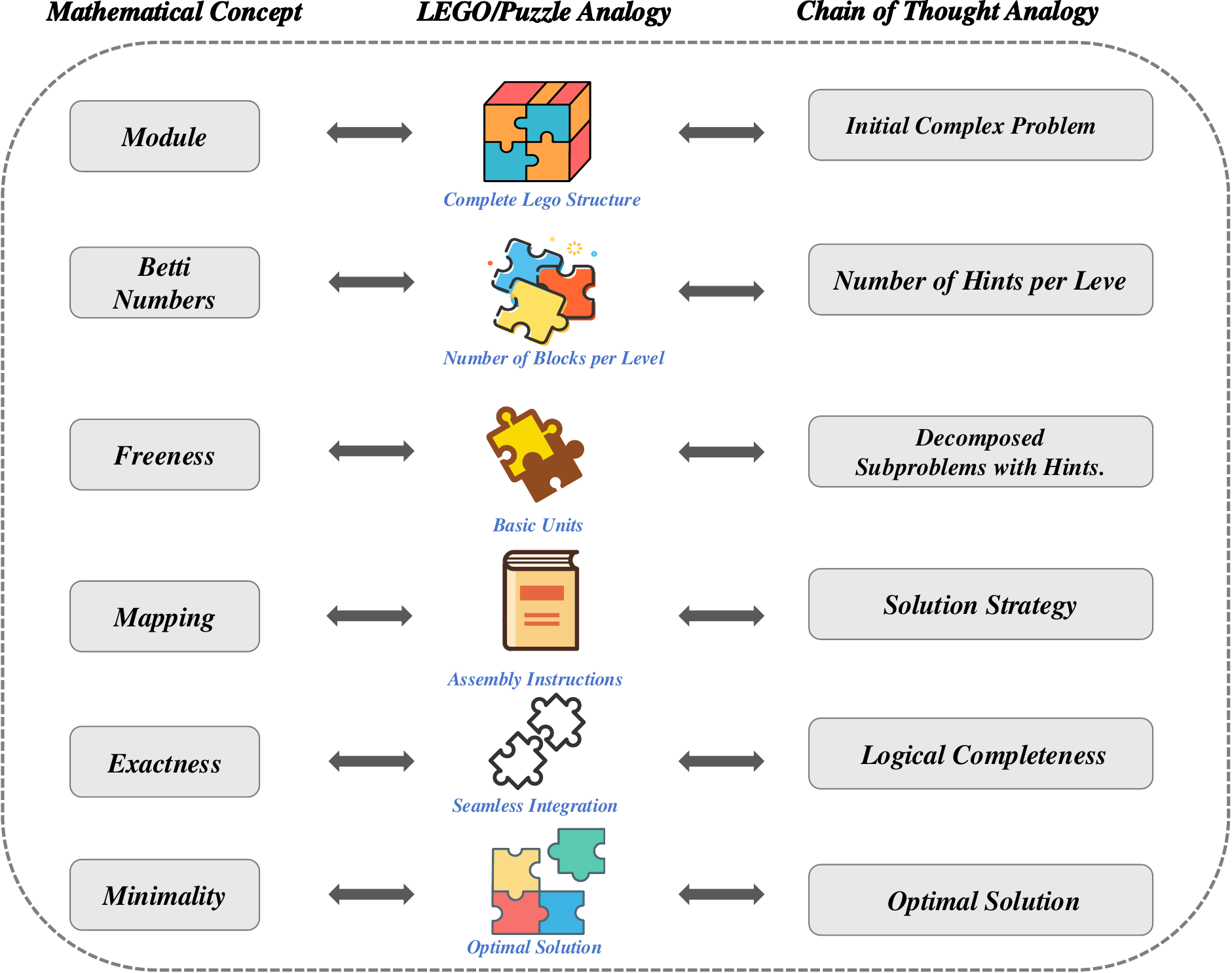}
    \caption{The mathematical, abstract analogy, and CoT analogy of Module, Freeness, Mapping, Exactness, Minimality, and Betti Number.}
    \label{fig:analogy}
\end{figure}

\subsection{Preliminaries}
\begin{figure*}
    \centering
    \includegraphics[width=\linewidth]{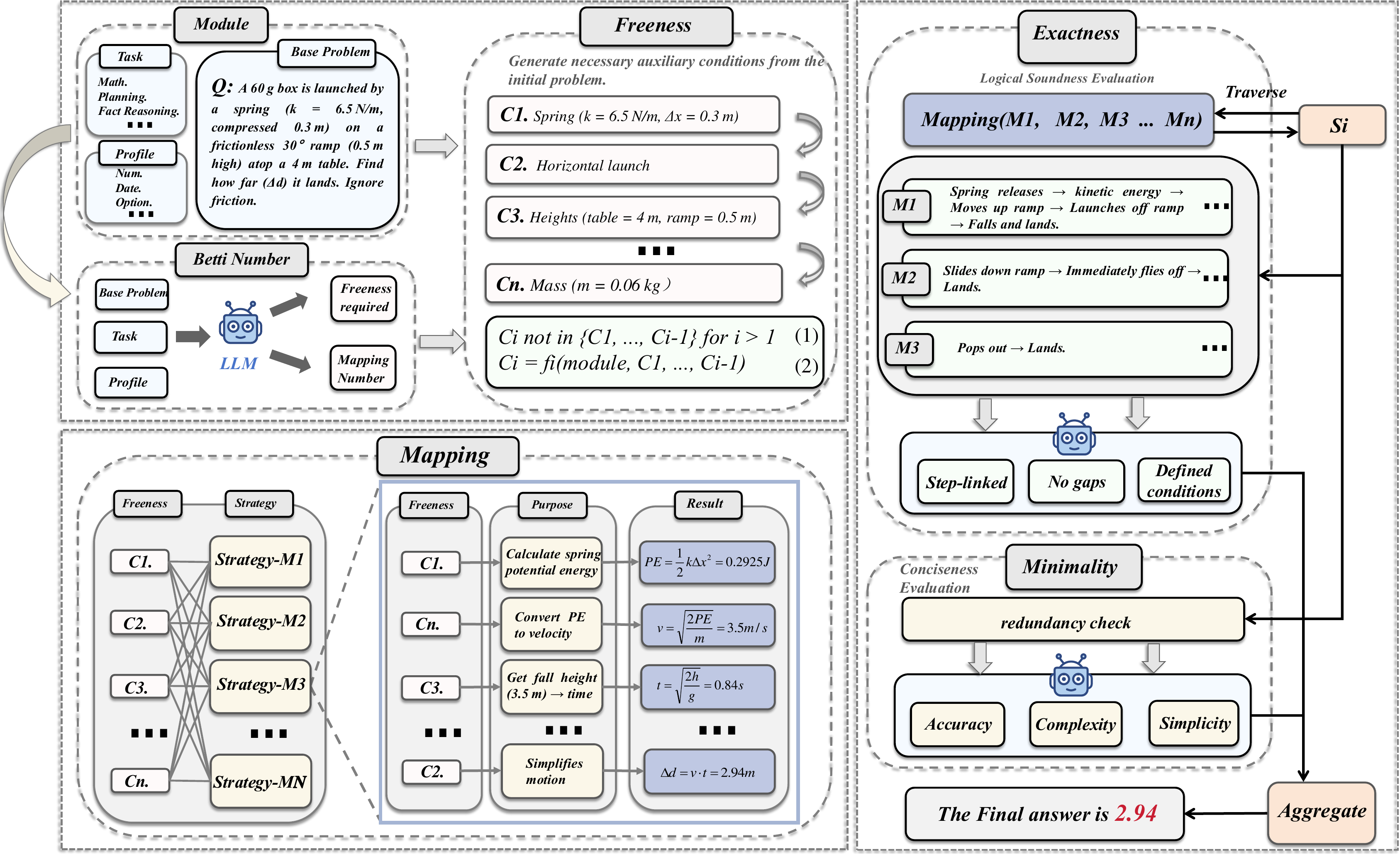}
    \caption{SoT Overview. Through the six modules: Module Freeness, Mapping, Exactness, Minimality, and Betti, MFR can decompose and deconstruct a complex reasoning problem, aiding LLMs in generating more accurate answers.}
    \label{fig:main}
\end{figure*}
In algebraic geometry and homological algebra, \textbf{Minimal Free Resolution} is a foundational tool for analyzing finitely generated modules over a ring, offering a structured way to reveal their internal dependencies and complexity.

\noindent\textbf{Key Concepts.} A module $M$ over a ring $R$ generalizes vector spaces by allowing scalars from $R$. $M$ is free if it admits a basis: $M \cong R^{\oplus n}$; it is finitely generated if a finite set $\{m_1, \dots, m_n\}$ spans $M$ via $Rm_i$.

A \textbf{syzygy} refers to a relation among generators, such as:
\begin{align}
    a_1 f_1 + a_2 f_2 + \dots + a_k f_k = 0,\quad a_i \in R,
\end{align}
and the collection of all such relations forms the \textit{first syzygy module}.

An MFR of $M$ is an exact sequence:
\begin{align}
    \cdots \to F_2 \to F_1 \to F_0 \to M \to 0,
\end{align}
where each $F_i$ is free and the maps contain no unit elements. The resolution is minimal if the number of generators in each $F_i$ is smallest possible, making the structure unique up to isomorphism.

The $i$-th Betti number, $\beta_i(M) = \text{rank}(F_i)$, quantifies the number of generators at each level and reflects the complexity of $M$.

\noindent\textbf{Motivation for SoT.} As illustrated in Figure~\ref{fig:main}, we reinterpret CoT reasoning within LLMs through the lens of MFR. Each reasoning stage—e.g., problem understanding, decomposition, subgoal chaining, and final decision—is treated as a generator or relation within a module. This enables SoT to capture latent logical dependencies and hierarchical structure, moving beyond CoT's linear stepwise form toward a more principled and algebraically grounded representation.

The SoT reasoning paradigm employs the MFR framework to model the structure of CoT reasoning paths in LLMs. Each reasoning step—such as problem interpretation, task decomposition, subgoal chaining, and final inference—is abstracted as generators or relations within modules, reinterpreting CoT as algebraic objects with generative structures and latent dependencies, rather than merely a flat sequence of symbols.

\noindent\includegraphics[width=0.03\textwidth]{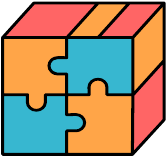}
\textit{Module: Initial Complex Problem}\\
Complex problems often exhibit high-dimensional structures, intricate logical dependencies, and ambiguous constraints, rendering direct solutions intractable. In this initial module, we conceptualize the problem as a unified entity of interdependent substructures, necessitating systematic decomposition. The goal is not immediate resolution but to unravel latent complexity, identify core logical components, and establish a foundation for subsequent modular reasoning. This stage ensures that downstream processes operate within well-defined boundaries, setting the stage for structured analysis.

\noindent\includegraphics[width=0.03\textwidth]{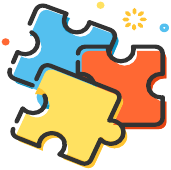}
\textit{Betti Numbers: Quantifying Complexity}\\
To measure decomposition complexity, we introduce Betti numbers as a quantitative metric, where each number reflects the count of auxiliary conditions at a reasoning level. Higher Betti numbers indicate greater intricacy, signaling potential optimization opportunities. We leverage LLMs to minimize these numbers by regenerating or filtering conditions, streamlining the decomposition process. Betti numbers thus guide reasoning efficiency, balancing structural complexity with computational simplicity.

\noindent\includegraphics[width=0.03\textwidth]{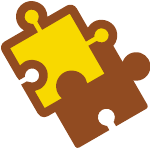}
\textit{Freeness}
\textit{: Auxiliary Conditions} \\
Freeness involves generating auxiliary conditions to simplify problems and clarify logical relationships. By introducing intermediate variables, hypotheses, or formalized constraints, complex problems are partitioned into smaller, independent subproblems. This reduces dimensionality and provides clear entry points for resolution, enhancing the tractability of subsequent steps while maintaining logical coherence.

\noindent\includegraphics[width=0.02\textwidth]{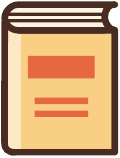}
\textit{Mapping: Solving Strategies}\\
Effective problem-solving requires systematic strategies that map auxiliary conditions to actionable reasoning paths. These paths must exhibit \textbf{(\textit{i})} directness, minimizing redundant operations, and \textbf{(\textit{ii})} logical soundness, grounding each step in explicit premises. LLMs iteratively solve subproblems along these paths, linking current conditions to new conclusions, ensuring continuity and interpretability throughout the reasoning chain.

\noindent\includegraphics[width=0.03\textwidth]{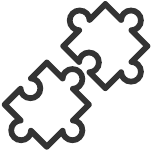}
\textit{Exactness: Logical Completeness}\\
Logical completeness is critical for rigor, preventing gaps or leaps in the reasoning chain. Each step must derive from explicit premises, with implicit assumptions formalized to avoid invalid conclusions. LLMs verify consistency across auxiliary conditions, subproblems, and the final solution, systematically reviewing the chain against original constraints. This ensures a coherent and reliable framework, minimizing errors from incomplete logic.

\noindent\includegraphics[width=0.027\textwidth]{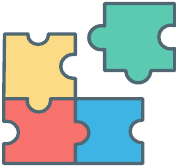}
\textit{Minimality: Optimal Solutions}\\
Minimality optimizes efficiency by deriving solutions with the fewest auxiliary conditions and simplest strategies, without sacrificing correctness. Redundant conditions are filtered, and reasoning paths are pruned of unnecessary steps. LLMs prioritize straightforward approaches over complex transformations, yielding concise, clear solutions. This reduces overall complexity, enhancing both computational efficiency and solution transparency.

After introducing the above six concepts, LLMs can be directly stimulated conversationally. Figure~\ref{fig:conceptual} illustrates the conversational framework of these six concepts. First, complex reasoning problems are defined as a free module. Then, through the Betti number and auxiliary conditions of degrees of freedom, the complex problem is reduced and decomposed. By analyzing both the module and degrees of freedom simultaneously, the model can construct multiple mappings to solve the problem. Finally, minimality and accuracy are used to select the optimal solution.

\subsection{Syzygy of Thoughts}

The process of LLM reasoning for complex problems can be viewed as solving a complex reasoning problem $M$ through MFR. The entire process is formalized via an MFR sequence, where the structured representation of the reasoning process is a set of mappings $\varphi_i$ induced by the LLM, forming the following sequence:

\begin{align}
\cdots 
\xrightarrow{\mathrm{LLM}(\varphi_n)} F_n 
\cdots 
\xrightarrow{\mathrm{LLM}(\varphi_0)} F_0 
\xrightarrow{} M \to 0
\end{align}

Moreover, we define the intermediate space $F_i$ as a free module constructed from basic reasoning units:

\begin{align}
    F_i = \bigoplus_j R \cdot m_{ij}
\end{align}

where $R$ is the reasoning space (LLM Reasoning Space), $m_{ij}$ are the basic units (Tokens) used in the reasoning construction. Each $\varphi_i$ is induced by the LLM, representing a single-step logical deduction in the CoT.

\subsubsection{SoT Reasoning.} Figure~\ref{fig:main} explains the entire process of SoT. The SoT framework begins with an initial complex problem $M$, characterized by high-dimensional structure, intricate logical dependencies, and ambiguous constraint relationships. We treat $M$ as a unified representation of interdependent substructures, setting the stage for systematic decomposition. To analyze the inherent complexity of $M$, we introduce Betti numbers as quantitative indicators of structural complexity. Specifically, the Betti number corresponds to the number of auxiliary conditions required for decomposition, with higher values suggesting deeper structural intricacy. These values guide the subsequent generation of auxiliary components and indicate potential optimization points.

We then introduce the notion of Freeness, which refers to the generation of auxiliary conditions that serve to simplify and structurally reorganize the problem. These conditions—denoted as $\Psi = \{\psi_1, \dots, \psi_\beta\}$—include intermediate hypotheses, latent constraints, or explicit sub-conclusions, and are generated via autoregressive reasoning using LLMs conditioned on previously established states. Each $\psi_j$ decomposes the original problem into finer-grained logical units, reducing its dimensional complexity and establishing solvable subspaces.

To resolve the decomposed problem, the framework constructs a set of mappings $\mathcal{M} = \{\mathcal{S}_1, \dots, \mathcal{S}_\mu\}$, each corresponding to a distinct reasoning path. These mappings are designed to satisfy two essential properties: directness, ensuring that each inference step avoids redundant operations, and logical soundness, ensuring that every conclusion is firmly grounded in its premises. Each mapping $\mathcal{S}_k$ is generated using LLMs to traverse from auxiliary conditions to intermediate or final conclusions, thereby forming a continuous and interpretable reasoning chain.

To ensure logical completeness (Exactness) of the reasoning process, all steps within each reasoning path are verified for inferential closure. Implicit assumptions are formalized, and the consistency of auxiliary conditions, mappings, and final conclusions is evaluated. This validation step leverages the LLM’s reasoning and scoring capabilities via a scoring function $\phi(\cdot)$. Finally, the optimal reasoning sequence $\mathcal{S}_{\text{opt}}$ is parsed using structured pattern-matching strategies (e.g., regular expressions, JSON parsing) to extract the final solution. 
\begin{table*}[t!]
    \centering
    \renewcommand{\arraystretch}{1}
    \scalebox{0.6}{
    \begin{tabular}{lccccccccc}
    \toprule
    \multirow{2}{*}{\textbf{Method}} &
    \multicolumn{5}{c}{\textbf{Math Reasoning}} &
    \multicolumn{1}{c}{\textbf{Gene.}} &
    \multicolumn{1}{c}{\textbf{Mul.}} &
    \multicolumn{1}{c}{\textbf{Tem.}} &
    \multicolumn{1}{c}{\textbf{Log.}} \\
    \cmidrule(lr){2-6} \cmidrule(lr){7-10}
     & \textbf{GSM8K} & \textbf{SVAMP} & \textbf{MultiArith} & \textbf{ASDiv} & \textbf{AQUA} &
     \textbf{MMLU} & \textbf{BBH} & \textbf{Date} & \textbf{CLUTRR} \\
    \midrule
    \multicolumn{10}{c}{\textbf{GPT-4o-mini}} \\
    \midrule
    CoT~\citep{wei2022chain} &
    85.4{\scriptsize$\pm 0.5$}\% &
    84.7{\scriptsize$\pm 1.6$}\% &
    89.5{\scriptsize$\pm 0.5$}\% &
    92.3{\scriptsize$\pm 0.8$}\% &
    65.3{\scriptsize$\pm 1.5$}\% &
    63.4{\scriptsize$\pm 0.5$}\% &
    66.6{\scriptsize$\pm 0.7$}\% &
    52.1{\scriptsize$\pm 1.0$}\% &
    66.2{\scriptsize$\pm 1.3$}\% \\
    CoT-SC~\citep{wang2022self} &
    89.9{\scriptsize$\pm 0.6$}\% &
    85.8{\scriptsize$\pm 0.8$}\% &
    89.7{\scriptsize$\pm 0.5$}\% &
    93.3{\scriptsize$\pm 1.0$}\% &
    70.7{\scriptsize$\pm 1.0$}\% &
    67.1{\scriptsize$\pm 0.5$}\% &
    69.0{\scriptsize$\pm 0.8$}\% &
    54.7{\scriptsize$\pm 0.7$}\% &
    72.2{\scriptsize$\pm 1.1$}\% \\
    GoT~\citep{besta2024graph} &
    93.2{\scriptsize$\pm 0.7$}\% & 
    89.2{\scriptsize$\pm 0.6$}\% &
    90.9{\scriptsize$\pm 0.5$}\% &
    94.2{\scriptsize$\pm 0.6$}\% &
    73.3{\scriptsize$\pm 1.1$}\% &
    71.3{\scriptsize$\pm 0.8$}\% &
    71.1{\scriptsize$\pm 0.9$}\% &
    65.2{\scriptsize$\pm 0.7$}\% &
    74.2{\scriptsize$\pm 1.0$}\% \\
    AoT~\citep{teng2025atom} &
    95.3{\scriptsize$\pm 0.5$}\% &
    91.5{\scriptsize$\pm 0.6$}\% &
    91.0{\scriptsize$\pm 0.5$}\% &
    94.1{\scriptsize$\pm 0.5$}\% &
    75.1{\scriptsize$\pm 1.0$}\% &
    74.7{\scriptsize$\pm 0.9$}\% &
    72.4{\scriptsize$\pm 0.7$}\% &
    74.7{\scriptsize$\pm 0.8$}\% &
    75.2{\scriptsize$\pm 1.2$}\% \\
    \textbf{SoT (Ours)} &
    \textbf{96.4{\scriptsize$\pm 0.6$}\%} &
    \textbf{92.6{\scriptsize$\pm 0.7$}\%} &
    \textbf{92.1{\scriptsize$\pm 0.7$}\%} &
    \textbf{95.2{\scriptsize$\pm 0.7$}\%} &
    \textbf{76.0{\scriptsize$\pm 1.1$}\%} &
    \textbf{75.6{\scriptsize$\pm 0.5$}\%} &
    \textbf{73.2{\scriptsize$\pm 0.6$}\%} &
    \textbf{75.6{\scriptsize$\pm 0.7$}\%} &
    \textbf{76.1{\scriptsize$\pm 0.7$}\%} \\
    \midrule
    \multicolumn{10}{c}{\textbf{Qwen2.5-Coder-7B-Instruct}} \\
    \midrule
    CoT~\citep{wei2022chain} &
    77.5{\scriptsize$\pm 1.4$}\% &
    82.7{\scriptsize$\pm 1.0$}\% &
    84.6{\scriptsize$\pm 0.9$}\% &
    87.3{\scriptsize$\pm 1.4$}\% &
    60.9{\scriptsize$\pm 1.2$}\% &
    55.4{\scriptsize$\pm 0.6$}\% &
    47.4{\scriptsize$\pm 0.6$}\% &
    31.3{\scriptsize$\pm 1.1$}\% &
    20.4{\scriptsize$\pm 1.1$}\% \\
    CoT-SC~\citep{wang2022self} &
    80.0{\scriptsize$\pm 0.9$}\% &
    83.9{\scriptsize$\pm 1.1$}\% &
    86.8{\scriptsize$\pm 0.6$}\% &
    89.8{\scriptsize$\pm 0.6$}\% &
    62.0{\scriptsize$\pm 1.4$}\% &
    56.1{\scriptsize$\pm 0.8$}\% &
    49.1{\scriptsize$\pm 0.9$}\% &
    32.7{\scriptsize$\pm 1.0$}\% &
    20.8{\scriptsize$\pm 1.0$}\% \\
    GoT~\citep{besta2024graph} &
    84.8{\scriptsize$\pm 0.7$}\% &
    87.5{\scriptsize$\pm 0.8$}\% &
    88.1{\scriptsize$\pm 0.5$}\% &
    92.5{\scriptsize$\pm 0.6$}\% &
    62.9{\scriptsize$\pm 1.2$}\% &
    56.8{\scriptsize$\pm 0.7$}\% &
    53.4{\scriptsize$\pm 0.7$}\% &
    34.7{\scriptsize$\pm 0.9$}\% &
    23.8{\scriptsize$\pm 1.0$}\% \\
    AoT~\citep{teng2025atom} &
    88.5{\scriptsize$\pm 0.6$}\% &
    90.0{\scriptsize$\pm 0.5$}\% &
    88.4{\scriptsize$\pm 0.7$}\% &
    94.1{\scriptsize$\pm 0.5$}\% &
    63.0{\scriptsize$\pm 1.0$}\% &
    56.8{\scriptsize$\pm 1.1$}\% &
    57.0{\scriptsize$\pm 1.0$}\% &
    36.2{\scriptsize$\pm 0.8$}\% &
    26.4{\scriptsize$\pm 1.3$}\% \\
    \textbf{SoT (Ours)} &
    \textbf{89.5{\scriptsize$\pm 0.6$}\%} &
    \textbf{91.0{\scriptsize$\pm 0.8$}\%} &
    \textbf{89.4{\scriptsize$\pm 0.6$}\%} &
    \textbf{95.2{\scriptsize$\pm 0.6$}\%} &
    \textbf{63.7{\scriptsize$\pm 1.0$}\%} &
    \textbf{57.5{\scriptsize$\pm 0.8$}\%} &
    \textbf{57.7{\scriptsize$\pm 0.7$}\%} &
    \textbf{36.6{\scriptsize$\pm 1.0$}\%} &
    \textbf{26.7{\scriptsize$\pm 1.1$}\%} \\
    \midrule
    \multicolumn{10}{c}{\textbf{Qwen2.5-VL-72B-Instruct}} \\
    \midrule
    CoT~\citep{wei2022chain} &
    86.4{\scriptsize$\pm 0.9$}\% &
    87.2{\scriptsize$\pm 0.9$}\% &
    90.1{\scriptsize$\pm 0.6$}\% &
    92.3{\scriptsize$\pm 0.6$}\% &
    81.4{\scriptsize$\pm 0.6$}\% &
    80.4{\scriptsize$\pm 0.5$}\% &
    77.6{\scriptsize$\pm 0.3$}\% &
    75.5{\scriptsize$\pm 0.5$}\% &
    70.4{\scriptsize$\pm 0.6$}\% \\
    CoT-SC~\citep{wang2022self} &
    88.9{\scriptsize$\pm 0.6$}\% &
    88.0{\scriptsize$\pm 0.6$}\% &
    91.1{\scriptsize$\pm 0.8$}\% &
    93.6{\scriptsize$\pm 0.6$}\% &
    83.7{\scriptsize$\pm 1.0$}\% &
    82.7{\scriptsize$\pm 0.5$}\% &
    78.8{\scriptsize$\pm 0.8$}\% &
    77.8{\scriptsize$\pm 0.8$}\% &
    74.8{\scriptsize$\pm 0.7$}\% \\
    GoT~\citep{besta2024graph} &
    92.7{\scriptsize$\pm 0.5$}\% &
    92.1{\scriptsize$\pm 0.4$}\% &
    91.6{\scriptsize$\pm 0.6$}\% &
    93.1{\scriptsize$\pm 0.5$}\% &
    86.8{\scriptsize$\pm 0.8$}\% &
    83.7{\scriptsize$\pm 0.6$}\% &
    82.2{\scriptsize$\pm 0.6$}\% &
    79.2{\scriptsize$\pm 0.7$}\% &
    77.0{\scriptsize$\pm 1.2$}\% \\
    AoT~\citep{teng2025atom} &
    95.3{\scriptsize$\pm 0.4$}\% &
    95.1{\scriptsize$\pm 0.5$}\% &
    91.0{\scriptsize$\pm 0.5$}\% &
    93.5{\scriptsize$\pm 0.5$}\% &
    88.8{\scriptsize$\pm 0.7$}\% &
    83.7{\scriptsize$\pm 0.7$}\% &
    84.7{\scriptsize$\pm 0.6$}\% &
    79.7{\scriptsize$\pm 0.8$}\% &
    78.4{\scriptsize$\pm 1.1$}\% \\
    \textbf{SoT (Ours)} &
    \textbf{96.4{\scriptsize$\pm 0.3$}\%} &
    \textbf{96.2{\scriptsize$\pm 0.2$}\%} &
    \textbf{92.1{\scriptsize$\pm 0.3$}\%} &
    \textbf{94.6{\scriptsize$\pm 0.5$}\%} &
    \textbf{89.8{\scriptsize$\pm 0.7$}\%} &
    \textbf{84.7{\scriptsize$\pm 0.8$}\%} &
    \textbf{85.7{\scriptsize$\pm 0.8$}\%} &
    \textbf{80.6{\scriptsize$\pm 1.5$}\%} &
    \textbf{79.3{\scriptsize$\pm 2.6$}\%} \\
    \midrule
    \multicolumn{10}{c}{\textbf{Gemma-3-27b-it}} \\
    \midrule
    CoT~\citep{wei2022chain} &
    83.4{\scriptsize$\pm 0.8$}\% &
    86.2{\scriptsize$\pm 0.9$}\% &
    84.2{\scriptsize$\pm 0.9$}\% &
    90.8{\scriptsize$\pm 0.4$}\% &
    80.6{\scriptsize$\pm 1.0$}\% &
    71.1{\scriptsize$\pm 1.6$}\% &
    71.0{\scriptsize$\pm 1.7$}\% &
    77.2{\scriptsize$\pm 1.4$}\% &
    65.6{\scriptsize$\pm 3.9$}\% \\
    CoT-SC~\citep{wang2022self} &
    86.9{\scriptsize$\pm 1.7$}\% &
    86.8{\scriptsize$\pm 1.8$}\% &
    84.5{\scriptsize$\pm 0.9$}\% &
    92.0{\scriptsize$\pm 0.5$}\% &
    85.2{\scriptsize$\pm 1.9$}\% &
    73.0{\scriptsize$\pm 3.5$}\% &
    73.0{\scriptsize$\pm 3.6$}\% &
    80.0{\scriptsize$\pm 2.7$}\% &
    66.2{\scriptsize$\pm 4.8$}\% \\
    GoT~\citep{besta2024graph} &
    91.7{\scriptsize$\pm 0.7$}\% &
    91.5{\scriptsize$\pm 0.6$}\% &
    88.3{\scriptsize$\pm 0.5$}\% &
    91.3{\scriptsize$\pm 0.5$}\% &
    87.5{\scriptsize$\pm 1.1$}\% &
    78.8{\scriptsize$\pm 1.0$}\% &
    79.3{\scriptsize$\pm 0.9$}\% &
    80.3{\scriptsize$\pm 0.8$}\% &
    72.8{\scriptsize$\pm 1.4$}\% \\
    AoT~\citep{teng2025atom} &
    93.2{\scriptsize$\pm 0.4$}\% &
    93.1{\scriptsize$\pm 0.4$}\% &
    88.4{\scriptsize$\pm 0.5$}\% &
    91.8{\scriptsize$\pm 0.6$}\% &
    85.6{\scriptsize$\pm 0.8$}\% &
    81.7{\scriptsize$\pm 0.8$}\% &
    82.9{\scriptsize$\pm 0.7$}\% &
    77.4{\scriptsize$\pm 0.7$}\% &
    76.1{\scriptsize$\pm 1.2$}\% \\
    \textbf{SoT (Ours)} &
    \textbf{94.2{\scriptsize$\pm 0.4$}\%} &
    \textbf{95.3{\scriptsize$\pm 0.2$}\%} &
    \textbf{91.9{\scriptsize$\pm 0.3$}\%} &
    \textbf{93.3{\scriptsize$\pm 0.4$}\%} &
    \textbf{88.1{\scriptsize$\pm 0.8$}\%} &
    \textbf{83.4{\scriptsize$\pm 1.4$}\%} &
    \textbf{84.7{\scriptsize$\pm 1.2$}\%} &
    79.6{\scriptsize$\pm 1.6$}\% &
    \textbf{78.4{\scriptsize$\pm 2.7$}\%} \\
    \midrule
    \multicolumn{10}{c}{\textbf{Gemma-3-12b-it}} \\
    \midrule
    CoT~\citep{wei2022chain} &
    83.5{\scriptsize$\pm 1.9$}\% &
    79.3{\scriptsize$\pm 2.0$}\% &
    82.7{\scriptsize$\pm 0.8$}\% &
    91.0{\scriptsize$\pm 0.4$}\% &
    69.2{\scriptsize$\pm 3.2$}\% &
    68.4{\scriptsize$\pm 2.7$}\% &
    64.9{\scriptsize$\pm 1.8$}\% &
    78.0{\scriptsize$\pm 1.9$}\% &
    49.3{\scriptsize$\pm 4.1$}\% \\
    CoT-SC~\citep{wang2022self} &
    85.9{\scriptsize$\pm 1.8$}\% &
    80.8{\scriptsize$\pm 0.9$}\% &
    85.1{\scriptsize$\pm 0.7$}\% &
    93.2{\scriptsize$\pm 0.4$}\% &
    71.5{\scriptsize$\pm 2.1$}\% &
    70.4{\scriptsize$\pm 1.6$}\% &
    66.5{\scriptsize$\pm 3.7$}\% &
    80.0{\scriptsize$\pm 1.8$}\% &
    52.0{\scriptsize$\pm 2.0$}\% \\
    GoT~\citep{besta2024graph} &
    89.2{\scriptsize$\pm 0.7$}\% &
    86.8{\scriptsize$\pm 0.6$}\% &
    86.8{\scriptsize$\pm 0.5$}\% &
    93.9{\scriptsize$\pm 0.6$}\% &
    74.5{\scriptsize$\pm 1.0$}\% &
    71.6{\scriptsize$\pm 1.1$}\% &
    68.0{\scriptsize$\pm 1.0$}\% &
    81.5{\scriptsize$\pm 0.8$}\% &
    53.7{\scriptsize$\pm 1.3$}\% \\
    AoT~\citep{teng2025atom} &
    91.4{\scriptsize$\pm 0.5$}\% &
    91.8{\scriptsize$\pm 0.4$}\% &
    87.5{\scriptsize$\pm 0.6$}\% &
    93.5{\scriptsize$\pm 0.5$}\% &
    76.7{\scriptsize$\pm 1.2$}\% &
    71.9{\scriptsize$\pm 1.3$}\% &
    68.7{\scriptsize$\pm 1.1$}\% &
    82.0{\scriptsize$\pm 0.9$}\% &
    54.8{\scriptsize$\pm 1.5$}\% \\
    \textbf{SoT (Ours)} &
    \textbf{92.5{\scriptsize$\pm 0.4$}\%} &
    \textbf{92.9{\scriptsize$\pm 0.5$}\%} &
    \textbf{88.5{\scriptsize$\pm 0.7$}\%} &
    \textbf{94.6{\scriptsize$\pm 0.3$}\%} &
    \textbf{77.6{\scriptsize$\pm 1.6$}\%} &
    \textbf{72.7{\scriptsize$\pm 1.5$}\%} &
    \textbf{69.5{\scriptsize$\pm 2.6$}\%} &
    \textbf{82.9{\scriptsize$\pm 0.7$}\%} &
    \textbf{55.4{\scriptsize$\pm 2.9$}\%} \\
    \bottomrule
\end{tabular}
    }
\caption{Performance comparison of CoT, CoT-SC ($n=5$), GoT ($b=4$, $L=3$), AoT ($depth<6$), and the proposed SoT method across various tasks, including mathematical reasoning, general knowledge (Gene.), multitask QA (Mul.), temporal reasoning (Tem.) and logical reasoning (Log.). We report mean accuracy $\pm$ standard deviation over 5 seeds. Results are grouped by model architectures, with SoT consistently achieving the best performance across benchmarks while maintaining comparable or lower variance.}
\label{tab:main}
\end{table*}

\section{Experiments}

\subsection{Applicability Evaluation}\label{sec:applicability}

\noindent\textbf{Configuration.} As shown in Table~\ref{tab:main}, we compare the proposed SoT method with four representative reasoning strategies: Chain-of-Thought (CoT)~\citep{wei2022chain}, Self-Consistency CoT (CoT-SC)~\citep{wang2022self}, Graph-of-Thought (GoT)~\citep{besta2024graph}, and Atom-of-Thought (AoT)~\citep{teng2025atom}. 
The evaluation covers nine benchmarks spanning four categories: mathematical word problems (GSM8K, SVAMP, MultiArith, ASDiv, AQUA), general-knowledge and multitask QA (MMLU, BBH), temporal reasoning (Date), and relational logical reasoning (CLUTRR). 
For CoT-SC, we follow the standard setting with $n=5$ sampled reasoning paths. 
GoT is instantiated with beam size $b=4$ and search depth $L=3$, while AoT uses a maximum search depth of $5$, consistent with the original works. 
We instantiate all methods on five backbone LLMs: GPT-4o-mini, Qwen2.5-Coder-7B-Instruct, Qwen2.5-VL-72B-Instruct, Gemma-3-27b-it, and Gemma-3-12b-it. 
For each backbone, we keep decoding hyper-parameters (e.g., temperature, top-$p$, maximum generation length) fixed across methods and only change the reasoning protocol. 
We report mean accuracy and standard deviation over multiple random seeds to account for sampling variance.

\noindent\textbf{Results.} From Table~\ref{tab:main}, SoT consistently achieves the best performance across all benchmarks and all backbone models. 
On the mathematical reasoning benchmarks, SoT improves over the strongest baseline AoT by around $1$ percentage point on GSM8K and SVAMP for every model family (e.g., on GPT-4o-mini, GSM8K accuracy increases from $95.3\%$ to $96.4\%$), while also yielding smaller but steady gains on MultiArith and ASDiv. 
On the more challenging AQUA dataset, SoT brings additional improvements of roughly $0.7$--$1.0$ points across different backbones, indicating that searching over structured thoughts helps the model better handle multi-step algebraic reasoning and eliminate distractor options. 
For general-knowledge and multitask reasoning, SoT further improves MMLU and BBH by approximately $0.7$--$1.0$ points over AoT on all models, with Qwen2.5-VL-72B and Gemma-3-27b-it achieving the highest absolute accuracies. 
Finally, on temporal and relational logical reasoning (Date and CLUTRR), SoT again attains the best results under every backbone, with gains of up to about $1$ point, suggesting that the structured search procedure is beneficial beyond pure arithmetic reasoning. 
Overall, SoT dominates all baselines in every setting in Table~\ref{tab:main} while maintaining comparable or even lower variance, demonstrating that it provides a stable and broadly applicable improvement to LLM reasoning.

\subsection{Cost Test}
\label{sec:cost_exp}

\begin{figure}[t]
    \centering
    \includegraphics[width=\linewidth]{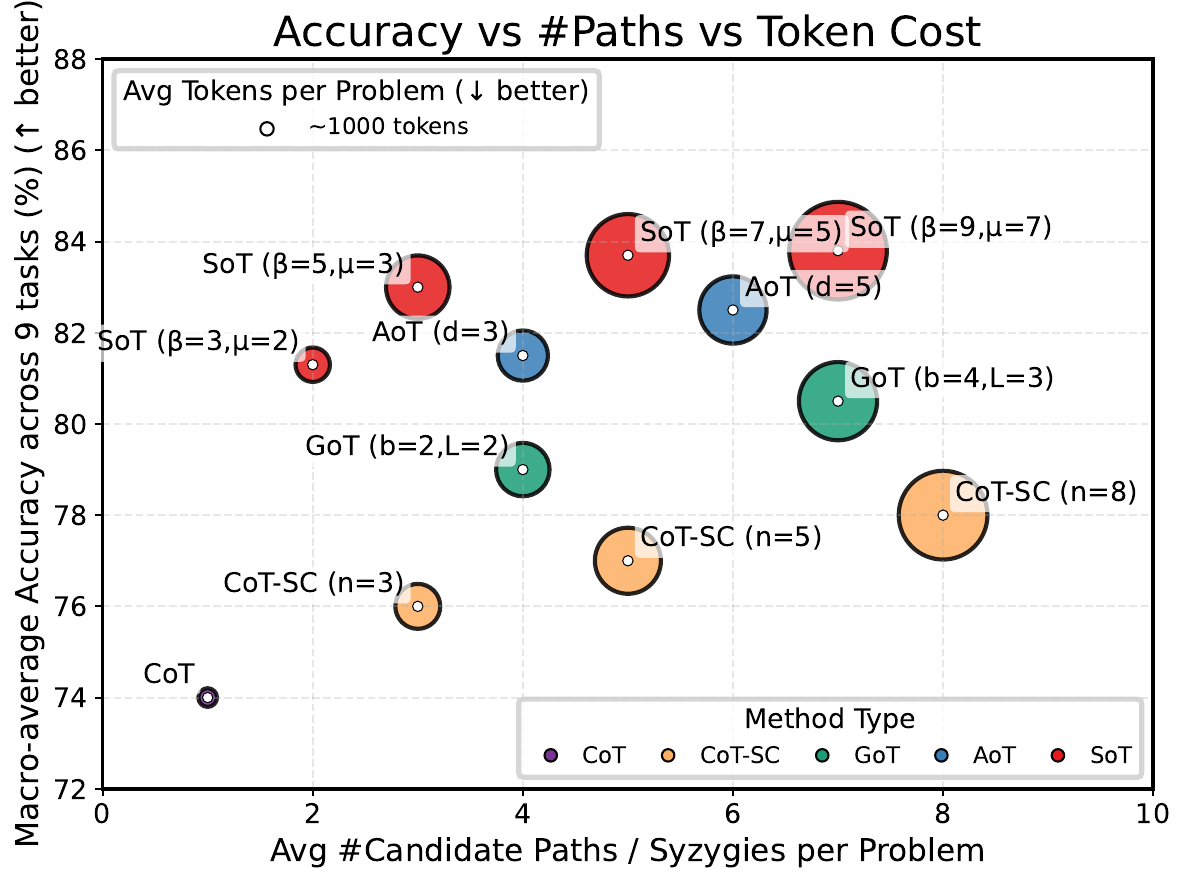}
    \caption{
        Cost–accuracy trade-off across different reasoning frameworks.
        Each bubble corresponds to one configuration of CoT, CoT-SC, GoT, AoT, or the proposed SoT, and the bubble area is proportional to the average number of tokens consumed per problem.
    }
    \label{fig:cost_bubble}
\end{figure}

To systematically compare the trade-offs between accuracy and search cost across different reasoning frameworks, we construct the three-dimensional bubble plot shown in Figure~\ref{fig:cost_bubble}, which places five methods under a unified lens: CoT, CoT-SC, GoT, AoT, and our proposed SoT. All methods are evaluated using the same model and the same set of nine benchmark tasks. In the plot, the vertical axis represents the macro-average accuracy over the nine tasks, the horizontal axis denotes the average number of candidate reasoning paths generated per problem, and the bubble area is proportional to the average number of tokens consumed per problem. We provide the detailed definition of our path-counting metric in Appendix~\ref{sec:cost_exp}.

Based on the above definitions, we visualize a representative set of method configurations. 
Along the horizontal axis, the number of candidate paths increases from left to right; 
along the vertical axis, the macro-average accuracy improves from bottom to top; 
and the bubble size grows with the token cost.
When the number of candidate paths and the token budget are comparable, 
SoT attains substantially higher macro-average accuracy than CoT-SC, GoT, and AoT. 
Moreover, as the number of syzygies increases from $\mu=3$ to $\mu=7$, 
the marginal gains diminish noticeably—indicating that SoT achieves near-saturated performance 
even with only a small number of syzygies.
Overall, SoT occupies a particularly favorable region in the three-way trade-off space: 
with a moderate number of candidate paths and an acceptable token cost, 
it achieves the highest macro-average accuracy and significantly enhances the 
effectiveness of each individual candidate path through its syzygy-based structure.

\subsection{Abalation Study}
\label{sec:ablation}

\begin{figure*}
    \centering
    \includegraphics[width=\linewidth]{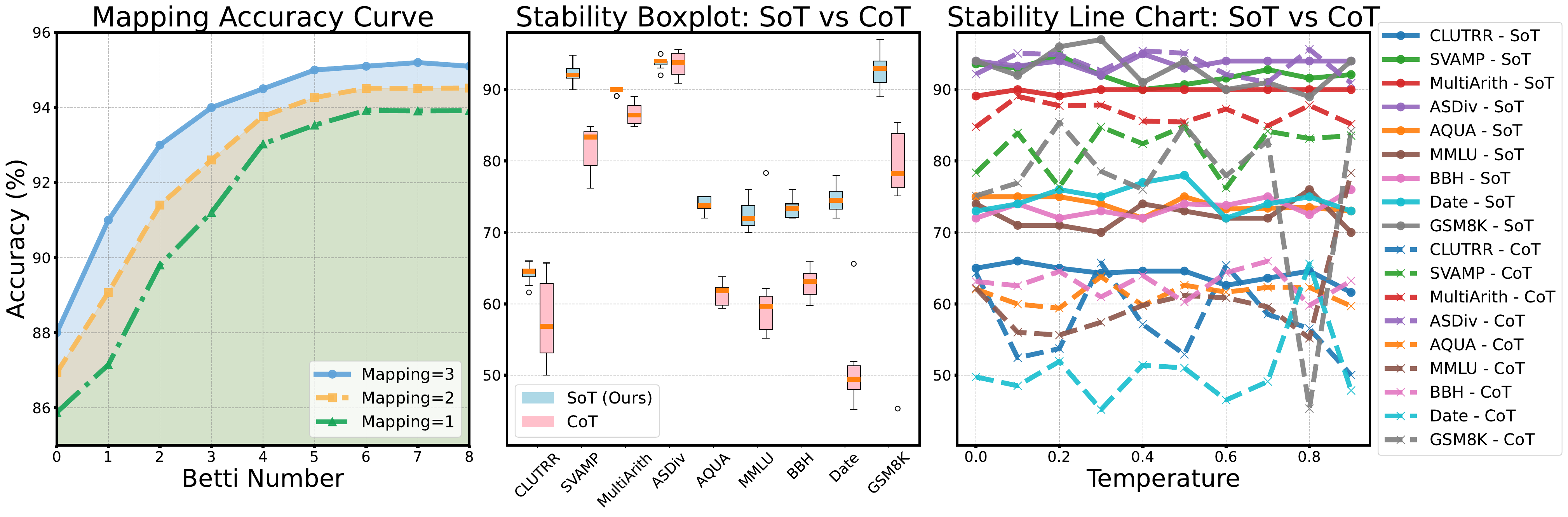}
    \caption{Overall analysis: (left) Betti number sensitivity, (middle) stability under different temperatures, and (right) accuracy distribution across tasks.}
    \label{fig:ablation}
\end{figure*}

\textbf{Betti Number Sensitivity.}
The SoT reasoning chain regulates topological constraints through the Betti number, and its value directly affects performance. To investigate the optimal configuration of the Betti number and its dynamic impact on structural expressiveness and regularization effects, we designed ablation experiments to reveal its key role in optimizing the reasoning process. For this, we systematically adjusted the Betti number and recorded accuracy changes under different mapping configurations. The goal of the experiment was to analyze the impact of the Betti number on the reasoning chain's expressiveness and identify the performance saturation point. The results are presented in Figure 5, showing the trend of accuracy changes with varying Betti numbers, with particular attention to its non-monotonic properties and optimal configuration.

\noindent\textit{Configuration.} The experiment was based on GPT-4o-mini and conducted on the GSM8K dataset. The Betti number was incremented from 1 to 10, with mapping set to 1, 2, and 3 to evaluate performance under different topological decomposition granularities.

\noindent\textit{Results.} As shown in Figure~\ref{fig:ablation} (left), accuracy changed non-monotonically as the Betti number increased under mapping configurations of 1, 2, and 3. Specifically, when the Betti number started from 1, accuracy improved significantly, indicating that a small number of topological constraints effectively enhanced structural expressiveness. However, as the Betti number increased further, the performance gain gradually diminished and stabilized. The optimal saturation point was reached at 7, beyond which there was no noticeable improvement. The results reveal the dynamic regulatory effect of the Betti number in reasoning chain modeling, and that moderate topological constraints are crucial for optimizing SoT’s structured design, providing strong support for its theoretical foundation.

\noindent\textbf{Stability Analysis.}
The temperature parameter influences the diversity of content generated by LLMs, which may challenge reasoning stability. We aimed to explore whether the SoT method could maintain its performance advantage under varying temperature conditions, thereby validating the adaptability of its structured design. To this end, we systematically evaluated SoT and CoT under different temperature conditions, focusing on the stability of accuracy and the relationship between reasoning diversity and consistency. The experiment involved adjusting the temperature step by step and observing the performance trends of the two methods under changing diversity.

\noindent\textit{Configuration.} The experiment was based on GPT-4o-mini, with the Betti number fixed at 7 and mapping set to 3. The nine test datasets included GSM8K, SVAMP, MultiArith, ASDiv, AQUA (mathematical reasoning), MMLU (generalization), BBH, Date (multi-task QA), and CLUTRR (logical reasoning). The temperature parameter was adjusted from 0.0 to 1.0 in steps of 0.1, and the performance of SoT and CoT was compared. Five reasoning paths were generated in each experiment, and the average accuracy was recorded to reduce the impact of randomness.

\noindent\textit{Results.} As shown in Figure~\ref{fig:ablation} (right), SoT's accuracy remains stable across temperature variations (0.0 to 1.0) with minimal fluctuation. In contrast, CoT's accuracy fluctuates significantly, especially at higher temperatures, where the amplitude increases. The boxplot in Figure~\ref{fig:ablation} (middle) reveals that SoT’s accuracy is concentrated with little variation and few outliers across datasets, even at high temperatures. On the other hand, CoT shows a wider spread with more outliers, especially at higher temperatures, where reasoning consistency decreases. This analysis indicates that CoT’s high temperature-induced diversity weakens logical coherence, while SoT maintains stability.
\section{Conclusions}
We introduced SoT, an MFR-inspired test-time reasoning framework that reframes CoT from a single linear trace into a syzygy-structured family of interdependent reasoning paths. By mapping MFR notions to concrete prompting and selection operations, SoT explicitly organizes auxiliary conditions, constructs candidate ``syzygies,'' verifies logical closure, and prunes redundancy to produce compact yet faithful solutions.
Across nine benchmarks spanning math, general knowledge, multitask QA, temporal reasoning, and relational logic, and across five LLM backbones, SoT consistently outperforms strong baselines while maintaining comparable or lower variance, indicating both improved accuracy and more stable inference. Moreover, SoT occupies a favorable region in the accuracy--cost--paths trade-off, achieving near-saturated gains with only a small number of syzygies, and exhibiting strong robustness under sampling diversity. 

\section{Limitations}
While SoT demonstrates strong reasoning capabilities, further research is needed. We plan to extend the topological decomposition to multimodal reasoning, including images and tables, to test its adaptability in cross-modal tasks. Additionally, incorporating iterative MFR concepts will refine problem-solving steps, optimize reasoning paths, and improve SoT's efficiency in multimodal and high-dimensional tasks.

\bibliography{custom}

\begin{thebibliography}{61}
\providecommand{\natexlab}[1]{#1}

\bibitem[{Besta et~al.(2024)Besta, Blach, Kubicek, Gerstenberger, Podstawski, Gianinazzi, Gajda, Lehmann, Niewiadomski, Nyczyk et~al.}]{besta2024graph}
Maciej Besta, Nils Blach, Ales Kubicek, Robert Gerstenberger, Michal Podstawski, Lukas Gianinazzi, Joanna Gajda, Tomasz Lehmann, Hubert Niewiadomski, Piotr Nyczyk, and 1 others. 2024.
\newblock Graph of thoughts: Solving elaborate problems with large language models.
\newblock In \emph{AAAI}.

\bibitem[{Botbol et~al.(2021)Botbol, Dickenstein, and Schenck}]{botbol2021simplest}
Nicol{\'a}s Botbol, Alicia Dickenstein, and Hal Schenck. 2021.
\newblock The simplest minimal free resolutions in {$\mathbb{P}^1$} $\times$ {$\mathbb{P}^1$}.
\newblock In \emph{Commutative Algebra: Expository Papers}.

\bibitem[{Brown et~al.(2020)Brown, Mann, Ryder, Subbiah, Kaplan, Dhariwal, Neelakantan, Shyam, Sastry, Askell et~al.}]{brown2020language}
Tom Brown, Benjamin Mann, Nick Ryder, Melanie Subbiah, Jared~D Kaplan, Prafulla Dhariwal, Arvind Neelakantan, Pranav Shyam, Girish Sastry, Amanda Askell, and 1 others. 2020.
\newblock Language models are few-shot learners.
\newblock In \emph{NeurIPS}.

\bibitem[{Capani et~al.(1997)Capani, De~Dominicis, Niesi, and Robbiano}]{capani1997computing}
Antonio Capani, Gabriel De~Dominicis, Gianfranco Niesi, and Lorenzo Robbiano. 1997.
\newblock Computing minimal finite free resolutions.
\newblock In \emph{J. Pure Appl. Algebra}.

\bibitem[{Chen et~al.(2025)Chen, Qin, Liu, Peng, Guan, Wang, Hu, Zhou, Gao, and Che}]{chen2025towards}
Qiguang Chen, Libo Qin, Jinhao Liu, Dengyun Peng, Jiannan Guan, Peng Wang, Mengkang Hu, Yuhang Zhou, Te~Gao, and Wangxiang Che. 2025.
\newblock Towards reasoning era: A survey of long chain-of-thought for reasoning large language models.
\newblock In \emph{arXiv}.

\bibitem[{Chen(2011)}]{chen2011hilbert}
Ri-Xiang Chen. 2011.
\newblock Hilbert functions and free resolutions.

\bibitem[{Chen et~al.(2022)Chen, Ma, Wang, and Cohen}]{chen2022program}
Wenhu Chen, Xueguang Ma, Xinyi Wang, and William~W Cohen. 2022.
\newblock Program of thoughts prompting: Disentangling computation from reasoning for numerical reasoning tasks.
\newblock In \emph{arXiv}.

\bibitem[{Chen et~al.(2024{\natexlab{a}})Chen, Chen, and Zhou}]{chen2024brain}
Yezeng Chen, Zui Chen, and Yi~Zhou. 2024{\natexlab{a}}.
\newblock Brain-inspired two-stage approach: Enhancing mathematical reasoning by imitating human thought processes.
\newblock In \emph{arXiv}.

\bibitem[{Chen et~al.(2024{\natexlab{b}})Chen, Deng, Yuan, Ji, and Gu}]{chen2024self}
Zixiang Chen, Yihe Deng, Huizhuo Yuan, Kaixuan Ji, and Quanquan Gu. 2024{\natexlab{b}}.
\newblock Self-play fine-tuning converts weak language models to strong language models.
\newblock In \emph{arXiv}.

\bibitem[{Cobbe et~al.(2021)Cobbe, Kosaraju, Bavarian, Chen, Jun, Kaiser, Plappert, Tworek, Hilton, Nakano et~al.}]{cobbe2021training}
Karl Cobbe, Vineet Kosaraju, Mohammad Bavarian, Mark Chen, Heewoo Jun, Lukasz Kaiser, Matthias Plappert, Jerry Tworek, Jacob Hilton, Reiichiro Nakano, and 1 others. 2021.
\newblock Training verifiers to solve math word problems.
\newblock In \emph{arXiv}.

\bibitem[{Deng et~al.(2024)Deng, Choi, and Shieber}]{deng2024explicit}
Yuntian Deng, Yejin Choi, and Stuart Shieber. 2024.
\newblock From explicit {CoT} to implicit {CoT}: Learning to internalize {CoT} step by step.
\newblock In \emph{arXiv}.

\bibitem[{Devlin et~al.(2019)Devlin, Chang, Lee, and Toutanova}]{devlin2019bert}
Jacob Devlin, Ming-Wei Chang, Kenton Lee, and Kristina Toutanova. 2019.
\newblock {BERT}: Pre-training of deep bidirectional transformers for language understanding.
\newblock In \emph{NAACL-HLT}.

\bibitem[{Eisenbud(2013)}]{eisenbud2013commutative}
David Eisenbud. 2013.
\newblock Commutative algebra: With a view toward algebraic geometry.
\newblock In \emph{Book}.

\bibitem[{Eisenbud et~al.(2001)Eisenbud, Grayson, Stillman, and Sturmfels}]{eisenbud2001computations}
David Eisenbud, Daniel~R Grayson, Mike Stillman, and Bernd Sturmfels. 2001.
\newblock Computations in algebraic geometry with {Macaulay} 2.
\newblock In \emph{Book}.

\bibitem[{Evans and Griffith(1981)}]{evans1981syzygy}
E~Graham Evans and Phillip Griffith. 1981.
\newblock The syzygy problem.
\newblock In \emph{Ann. Math.}

\bibitem[{Fieldsteel and Nagel(2021)}]{fieldsteel2021minimal}
Nathan Fieldsteel and Uwe Nagel. 2021.
\newblock Minimal and cellular free resolutions over polynomial {OI}-algebras.
\newblock In \emph{arXiv}.

\bibitem[{Gou et~al.(2023)Gou, Shao, Gong, Shen, Yang, Huang, Duan, and Chen}]{gou2023tora}
Zhibin Gou, Zhihong Shao, Yeyun Gong, Yelong Shen, Yujiu Yang, Minlie Huang, Nan Duan, and Weizhu Chen. 2023.
\newblock {ToRA}: A tool-integrated reasoning agent for mathematical problem solving.
\newblock In \emph{arXiv}.

\bibitem[{Hendrycks et~al.(2021)Hendrycks, Burns, Kadavath, Arora, Basart, Tang, Song, and Steinhardt}]{hendrycksmath2021}
Dan Hendrycks, Collin Burns, Saurav Kadavath, Akul Arora, Steven Basart, Eric Tang, Dawn Song, and Jacob Steinhardt. 2021.
\newblock Measuring mathematical problem solving with the {MATH} dataset.
\newblock In \emph{NeurIPS}.

\bibitem[{Hitchin(2003)}]{hitchin2003generalized}
Nigel Hitchin. 2003.
\newblock Generalized calabi--yau manifolds.
\newblock In \emph{Q. J. Math.}

\bibitem[{Imani et~al.(2023)Imani, Du, and Shrivastava}]{imani2023mathprompter}
Shima Imani, Liang Du, and Harsh Shrivastava. 2023.
\newblock {MathPrompter}: Mathematical reasoning using large language models.
\newblock In \emph{arXiv}.

\bibitem[{Jin et~al.(2024)Jin, Yu, Shu, Zhao, Hua, Meng, Zhang, and Du}]{jin2024impact}
Mingyu Jin, Qinkai Yu, Dong Shu, Haiyan Zhao, Wenyue Hua, Yanda Meng, Yongfeng Zhang, and Mengnan Du. 2024.
\newblock The impact of reasoning step length on large language models.
\newblock In \emph{arXiv}.

\bibitem[{Kojima et~al.(2022)Kojima, Gu, Reid, Matsuo, and Iwasawa}]{kojima2022large}
Takeshi Kojima, Shixiang~Shane Gu, Machel Reid, Yutaka Matsuo, and Yusuke Iwasawa. 2022.
\newblock Large language models are zero-shot reasoners.
\newblock In \emph{NeurIPS}.

\bibitem[{Leang et~al.(2024)Leang, Gema, and Cohen}]{leang2024comat}
Joshua Ong~Jun Leang, Aryo~Pradipta Gema, and Shay~B Cohen. 2024.
\newblock {CoMAT}: Chain of mathematically annotated thought improves mathematical reasoning.
\newblock In \emph{arXiv}.

\bibitem[{Li et~al.(2023)Li, Liang, Zeng, Chen, Hausman, Sadigh, Levine, Fei-Fei, Xia, and Ichter}]{li2023chain}
Chengshu Li, Jacky Liang, Andy Zeng, Xinyun Chen, Karol Hausman, Dorsa Sadigh, Sergey Levine, Li~Fei-Fei, Fei Xia, and Brian Ichter. 2023.
\newblock Chain of code: Reasoning with a language model-augmented code emulator.
\newblock In \emph{arXiv}.

\bibitem[{Li et~al.(2025)Li, Guo, Yang, Xu, Wu, and He}]{li2025codei}
Junlong Li, Daya Guo, Dejian Yang, Runxin Xu, Yu~Wu, and Junxian He. 2025.
\newblock {CodeI/O}: Condensing reasoning patterns via code input-output prediction.
\newblock In \emph{arXiv}.

\bibitem[{Li et~al.(2011)Li, Mou, Niu, and Wang}]{li2011stability}
Xiaoliang Li, Chenqi Mou, Wei Niu, and Dongming Wang. 2011.
\newblock Stability analysis for discrete biological models using algebraic methods.
\newblock In \emph{Math. Comput. Sci.}

\bibitem[{Lyu et~al.(2023)Lyu, Havaldar, Stein, Zhang, Rao, Wong, Apidianaki, and Callison-Burch}]{lyu2023faithful}
Qing Lyu, Shreya Havaldar, Adam Stein, Li~Zhang, Delip Rao, Eric Wong, Marianna Apidianaki, and Chris Callison-Burch. 2023.
\newblock Faithful chain-of-thought reasoning.
\newblock In \emph{IJCNLP-AACL}.

\bibitem[{Madaan et~al.(2023)Madaan, Tandon, Gupta, Hallinan, Gao, Wiegreffe, Alon, Dziri, Prabhumoye, Yang et~al.}]{madaan2023self}
Aman Madaan, Niket Tandon, Prakhar Gupta, Skyler Hallinan, Luyu Gao, Sarah Wiegreffe, Uri Alon, Nouha Dziri, Shrimai Prabhumoye, Yiming Yang, and 1 others. 2023.
\newblock Self-refine: Iterative refinement with self-feedback.
\newblock In \emph{NeurIPS}.

\bibitem[{Min et~al.(2023)Min, Ross, Sulem, Veyseh, Nguyen, Sainz, Agirre, Heintz, and Roth}]{min2023recent}
Bonan Min, Hayley Ross, Elior Sulem, Amir Pouran~Ben Veyseh, Thien~Huu Nguyen, Oscar Sainz, Eneko Agirre, Ilana Heintz, and Dan Roth. 2023.
\newblock Recent advances in natural language processing via large pre-trained language models: A survey.
\newblock In \emph{ACM Comput. Surv.}

\bibitem[{Nguyen et~al.(2023)Nguyen, Liu, Zhang, Zhang, and Yu}]{nguyen2023cof}
Hoang~H Nguyen, Ye~Liu, Chenwei Zhang, Tao Zhang, and Philip~S Yu. 2023.
\newblock Cof-cot: Enhancing large language models with coarse-to-fine chain-of-thought prompting for multi-domain nlu tasks.
\newblock In \emph{arXiv}.

\bibitem[{Ning et~al.(2023)Ning, Lin, Zhou, Wang, Yang, and Wang}]{ning2023skeleton}
Xuefei Ning, Zinan Lin, Zixuan Zhou, Zifu Wang, Huazhong Yang, and Yu~Wang. 2023.
\newblock Skeleton-of-thought: Large language models can do parallel decoding.
\newblock In \emph{ENLSP-III}.

\bibitem[{Polchinski(1994)}]{polchinski1994string}
Joseph Polchinski. 1994.
\newblock What is string theory?
\newblock In \emph{arXiv}.

\bibitem[{Radford et~al.(2018)Radford, Narasimhan, Salimans, Sutskever et~al.}]{radford2018improving}
Alec Radford, Karthik Narasimhan, Tim Salimans, Ilya Sutskever, and 1 others. 2018.
\newblock Improving language understanding by generative pre-training.
\newblock In \emph{arXiv}.

\bibitem[{Ranaldi et~al.(2025)Ranaldi, Valentino, Polonsky, and Freitas}]{ranaldi2025improving}
Leonardo Ranaldi, Marco Valentino, Alexander Polonsky, and Andr{\`e} Freitas. 2025.
\newblock Improving chain-of-thought reasoning via quasi-symbolic abstractions.
\newblock In \emph{arXiv}.

\bibitem[{Rossi and Sharifan(2009)}]{rossi2009minimal}
Maria~Evelina Rossi and Leila Sharifan. 2009.
\newblock Minimal free resolution of a finitely generated module over a regular local ring.
\newblock In \emph{J. Algebra}.

\bibitem[{Sel et~al.(2023)Sel, Al-Tawaha, Khattar, Jia, and Jin}]{sel2023algorithm}
Bilgehan Sel, Ahmad Al-Tawaha, Vanshaj Khattar, Ruoxi Jia, and Ming Jin. 2023.
\newblock Algorithm of thoughts: Enhancing exploration of ideas in large language models.
\newblock In \emph{arXiv}.

\bibitem[{Shi et~al.(2023)Shi, Chen, Misra, Scales, Dohan, Chi, Sch{\"a}rli, and Zhou}]{shi2023large}
Freda Shi, Xinyun Chen, Kanishka Misra, Nathan Scales, David Dohan, Ed~H Chi, Nathanael Sch{\"a}rli, and Denny Zhou. 2023.
\newblock Large language models can be easily distracted by irrelevant context.
\newblock In \emph{ICML}.

\bibitem[{Srivastava and Gandhi(2024)}]{srivastava2024mathdivide}
Saksham~Sahai Srivastava and Ashutosh Gandhi. 2024.
\newblock {MathDivide}: Improved mathematical reasoning by large language models.
\newblock In \emph{arXiv}.

\bibitem[{Stewart(1993)}]{stewart1993early}
Gilbert~W Stewart. 1993.
\newblock On the early history of the singular value decomposition.
\newblock In \emph{SIAM Rev.}

\bibitem[{Teng et~al.(2025)Teng, Yu, Shi, Zhang, Wu, and Luo}]{teng2025atom}
Fengwei Teng, Zhaoyang Yu, Quan Shi, Jiayi Zhang, Chenglin Wu, and Yuyu Luo. 2025.
\newblock Atom of thoughts for {Markov} {LLM} test-time scaling.
\newblock In \emph{arXiv}.

\bibitem[{Vaswani et~al.(2017)Vaswani, Shazeer, Parmar, Uszkoreit, Jones, Gomez, Kaiser, and Polosukhin}]{vaswani2017attention}
Ashish Vaswani, Noam Shazeer, Niki Parmar, Jakob Uszkoreit, Llion Jones, Aidan~N Gomez, {\L}ukasz Kaiser, and Illia Polosukhin. 2017.
\newblock Attention is all you need.
\newblock In \emph{NeurIPS}.

\bibitem[{Wang et~al.(2022{\natexlab{a}})Wang, Min, Deng, Shen, Wu, Zettlemoyer, and Sun}]{wang2022towards}
Boshi Wang, Sewon Min, Xiang Deng, Jiaming Shen, You Wu, Luke Zettlemoyer, and Huan Sun. 2022{\natexlab{a}}.
\newblock Towards understanding chain-of-thought prompting: An empirical study of what matters.
\newblock In \emph{arXiv}.

\bibitem[{Wang et~al.(2023{\natexlab{a}})Wang, Ren, Zhou, Lu, Luo, Shi, Zhang, Song, Zhan, and Li}]{wang2023mathcoder}
Ke~Wang, Houxing Ren, Aojun Zhou, Zimu Lu, Sichun Luo, Weikang Shi, Renrui Zhang, Linqi Song, Mingjie Zhan, and Hongsheng Li. 2023{\natexlab{a}}.
\newblock {MathCoder}: Seamless code integration in {LLMs} for enhanced mathematical reasoning.
\newblock In \emph{arXiv}.

\bibitem[{Wang et~al.(2023{\natexlab{b}})Wang, Caccia, Ostapenko, Yuan, Wang, and Sordoni}]{wang2023guiding}
Xinyi Wang, Lucas Caccia, Oleksiy Ostapenko, Xingdi Yuan, William~Yang Wang, and Alessandro Sordoni. 2023{\natexlab{b}}.
\newblock Guiding language model reasoning with planning tokens.
\newblock In \emph{arXiv}.

\bibitem[{Wang et~al.(2022{\natexlab{b}})Wang, Wei, Schuurmans, Le, Chi, Narang, Chowdhery, and Zhou}]{wang2022self}
Xuezhi Wang, Jason Wei, Dale Schuurmans, Quoc Le, Ed~Chi, Sharan Narang, Aakanksha Chowdhery, and Denny Zhou. 2022{\natexlab{b}}.
\newblock Self-consistency improves chain of thought reasoning in language models.
\newblock In \emph{arXiv}.

\bibitem[{Wasserman(2018)}]{wasserman2018topological}
Larry Wasserman. 2018.
\newblock Topological data analysis.
\newblock In \emph{Annu. Rev. Stat. Appl.}

\bibitem[{Wei et~al.(2022)Wei, Wang, Schuurmans, Bosma, Xia, Chi, Le, Zhou et~al.}]{wei2022chain}
Jason Wei, Xuezhi Wang, Dale Schuurmans, Maarten Bosma, Fei Xia, Ed~Chi, Quoc~V Le, Denny Zhou, and 1 others. 2022.
\newblock Chain-of-thought prompting elicits reasoning in large language models.
\newblock In \emph{NeurIPS}.

\bibitem[{Weispfenning(1992)}]{weispfenning1992comprehensive}
Volker Weispfenning. 1992.
\newblock Comprehensive gr{\"o}bner bases.
\newblock In \emph{J. Symb. Comput.}

\bibitem[{Xu et~al.(2024)Xu, Sun, Cheng, Liu, Qiao, and Wu}]{xu2024interactive}
Fangzhi Xu, Qiushi Sun, Kanzhi Cheng, Jun Liu, Yu~Qiao, and Zhiyong Wu. 2024.
\newblock Interactive evolution: A neural-symbolic self-training framework for large language models.
\newblock In \emph{arXiv}.

\bibitem[{Yang et~al.(2025)Yang, Yu, Zhang, Cao, Xu, Zhang, Gonzalez, and Cui}]{yang2025buffer}
Ling Yang, Zhaochen Yu, Tianjun Zhang, Shiyi Cao, Minkai Xu, Wentao Zhang, Joseph~E Gonzalez, and Bin Cui. 2025.
\newblock Buffer of thoughts: Thought-augmented reasoning with large language models.
\newblock In \emph{NeurIPS}.

\bibitem[{Yao et~al.(2023)Yao, Yu, Zhao, Shafran, Griffiths, Cao, and Narasimhan}]{yao2023tree}
Shunyu Yao, Dian Yu, Jeffrey Zhao, Izhak Shafran, Tom Griffiths, Yuan Cao, and Karthik Narasimhan. 2023.
\newblock Tree of thoughts: Deliberate problem solving with large language models.
\newblock In \emph{NeurIPS}.

\bibitem[{Yu et~al.(2024)Yu, Peng, Tian, Song, Mi, and Yu}]{yu2024siam}
Dian Yu, Baolin Peng, Ye~Tian, Linfeng Song, Haitao Mi, and Dong Yu. 2024.
\newblock {SIAM}: Self-improving code-assisted mathematical reasoning of large language models.
\newblock In \emph{arXiv}.

\bibitem[{Yu et~al.(2023{\natexlab{a}})Yu, Gao, and Wang}]{yu2023ovm}
Fei Yu, Anningzhe Gao, and Benyou Wang. 2023{\natexlab{a}}.
\newblock {OVM}: Outcome-supervised value models for planning in mathematical reasoning.
\newblock In \emph{arXiv}.

\bibitem[{Yu et~al.(2023{\natexlab{b}})Yu, Jiang, Shi, Yu, Liu, Zhang, Kwok, Li, Weller, and Liu}]{yu2023metamath}
Longhui Yu, Weisen Jiang, Han Shi, Jincheng Yu, Zhengying Liu, Yu~Zhang, James~T Kwok, Zhenguo Li, Adrian Weller, and Weiyang Liu. 2023{\natexlab{b}}.
\newblock {MetaMath}: Bootstrap your own mathematical questions for large language models.
\newblock In \emph{arXiv}.

\bibitem[{Yue et~al.(2023)Yue, Qu, Zhang, Fu, Huang, Sun, Su, and Chen}]{yue2023mammoth}
Xiang Yue, Xingwei Qu, Ge~Zhang, Yao Fu, Wenhao Huang, Huan Sun, Yu~Su, and Wenhu Chen. 2023.
\newblock {MAmmoTH}: Building math generalist models through hybrid instruction tuning.
\newblock In \emph{arXiv}.

\bibitem[{Zelikman et~al.(2022)Zelikman, Wu, Mu, and Goodman}]{zelikman2022star}
Eric Zelikman, Yuhuai Wu, Jesse Mu, and Noah Goodman. 2022.
\newblock {STAR}: Bootstrapping reasoning with reasoning.
\newblock In \emph{NeurIPS}.

\bibitem[{Zhang et~al.(2022)Zhang, Zhang, Li, and Smola}]{zhang2022automatic}
Zhuosheng Zhang, Aston Zhang, Mu~Li, and Alex Smola. 2022.
\newblock Automatic chain of thought prompting in large language models.
\newblock In \emph{arXiv}.

\bibitem[{Zhao et~al.(2023{\natexlab{a}})Zhao, Li, Joty, Qin, and Bing}]{zhao2023verify}
Ruochen Zhao, Xingxuan Li, Shafiq Joty, Chengwei Qin, and Lidong Bing. 2023{\natexlab{a}}.
\newblock Verify-and-edit: A knowledge-enhanced chain-of-thought framework.
\newblock In \emph{arXiv}.

\bibitem[{Zhao et~al.(2023{\natexlab{b}})Zhao, Zhou, Li, Tang, Wang, Hou, Min, Zhang, Zhang, Dong et~al.}]{zhao2023survey}
Wayne~Xin Zhao, Kun Zhou, Junyi Li, Tianyi Tang, Xiaolei Wang, Yupeng Hou, Yingqian Min, Beichen Zhang, Junjie Zhang, Zican Dong, and 1 others. 2023{\natexlab{b}}.
\newblock A survey of large language models.
\newblock In \emph{arXiv}.

\bibitem[{Zheng et~al.(2023)Zheng, Mishra, Chen, Cheng, Chi, Le, and Zhou}]{zheng2023take}
Huaixiu~Steven Zheng, Swaroop Mishra, Xinyun Chen, Heng-Tze Cheng, Ed~H Chi, Quoc~V Le, and Denny Zhou. 2023.
\newblock Take a step back: Evoking reasoning via abstraction in large language models.
\newblock In \emph{arXiv}.

\bibitem[{Zhou et~al.(2022)Zhou, Sch{\"a}rli, Hou, Wei, Scales, Wang, Schuurmans, Cui, Bousquet, Le et~al.}]{zhou2022least}
Denny Zhou, Nathanael Sch{\"a}rli, Le~Hou, Jason Wei, Nathan Scales, Xuezhi Wang, Dale Schuurmans, Claire Cui, Olivier Bousquet, Quoc Le, and 1 others. 2022.
\newblock Least-to-most prompting enables complex reasoning in large language models.
\newblock In \emph{arXiv}.

\end{thebibliography}
\newpage
\appendix
\onecolumn
\section{Appendix}
\label{sec:appendix}

\subsection{Full Pseudocode of the SoT Pipeline}

\begin{algorithm}[t]
\caption{LLM Minimal Free Resolution (LLM-MFR) for Reasoning Module $M$}
\label{alg:minimal_free_resolution_standard}
\begin{algorithmic}[1]
    \Require Reasoning question $M$; metadata $D$
    \Ensure Final answer $\text{Answer}$

    \State $(\beta, \mu) \gets \textsc{Analyse}(M, D)$
    \Comment{Estimate regularity $\beta$ and projective dimension $\mu$ (Sec.~\ref{sec:method})}

    \State \Comment{Generate auxiliary conditions (Freeness)}
    \State $\Psi \gets [\,]$ 
    \Comment{Initialize empty list of auxiliary conditions}
    \For{$j \gets 1$ \textbf{to} $\beta$}
        \State $\psi_j \gets \textsc{LLMGenerateAuxCondition}(F_0, \dots, F_{j-1}, \varphi_0, \dots, \varphi_{j-1})$
        \Comment{Use LLM to generate the $j$-th freeness condition}
        \State $\Psi \gets \Psi \cup \{\psi_j\}$
    \EndFor

    \State \Comment{Construct candidate resolutions (syzygies)}
    \State $\mathcal{M} \gets [\,]$
    \Comment{Initialize empty list of candidate solutions}
    \For{$k \gets 1$ \textbf{to} $\mu$}
        \State $\mathcal{S}_k \gets \textsc{LLMResolve}(M, \Psi)$
        \Comment{LLM resolves $M$ under auxiliary conditions $\Psi$}
        \State $\mathcal{M} \gets \mathcal{M} \cup \{\mathcal{S}_k\}$
    \EndFor

    \State \Comment{Select optimal resolution from syzygies (Minimality)}
    \State $k^* \gets \textsc{ArgMinScore}(\mathcal{M}, \phi)$
    \Comment{$\phi$: scoring function implemented via LLM scoring prompt}
    \State $\mathcal{S}_{\text{opt}} \gets \mathcal{S}_{k^*}$

    \State $\text{Answer} \gets \textsc{ExtractFinalAnswer}(\mathcal{S}_{\text{opt}}, D)$
    \Comment{Ensure logical Exactness via pattern matching (e.g., regex / JSON parsing)}

    \State \Return $\text{Answer}$
\end{algorithmic}
\end{algorithm}

In this appendix, we present the full pseudocode of the SoT pipeline.
Algorithm~\ref{alg:minimal_free_resolution_standard} provides an algorithmic implementation of the LLM-based minimal free resolution discussed in Section~\ref{sec:method}, turning the conceptual modules in Figure~\ref{fig:main} and the abstract diagram in Figure~\ref{fig:conceptual} into an executable procedure.

\paragraph{Explanation of Algorithm~\ref{alg:minimal_free_resolution_standard}.}
We briefly explain the main components of Algorithm~\ref{alg:minimal_free_resolution_standard}, which instantiates the SoT framework described in Section~\ref{sec:method}.
Analyse $(M, D)$ and homological complexity.  
Given a reasoning problem $M$ and its metadata $D$, the subroutine \textsc{Analyse}$(M, D)$ estimates the regularity $\beta$ and the projective dimension $\mu$. These quantities summarize the Betti-number profile introduced in Section~\ref{sec:method}: $\beta$ determines the number of auxiliary (freeness) conditions, and $\mu$ specifies how many candidate resolutions (syzygies) are explored. Larger values of $(\beta, \mu)$ indicate higher problem complexity.
Generating auxiliary conditions (Freeness).  
The loop over $j = 1, \dots, \beta$ constructs a set $\Psi$ of auxiliary conditions, corresponding to the ``Freeness'' module in Figure~\ref{fig:main}. Each call to \textsc{LLMGenerateAuxCondition} conditions on previously created free modules $F_0, \dots, F_{j-1}$ and maps $\varphi_0, \dots, \varphi_{j-1}$ to propose a new constraint $\psi_j$. These constraints guide the LLM toward structurally coherent reasoning decompositions.
Constructing candidate syzygies (Syzygy of Thoughts).  
The loop over $k = 1, \dots, \mu$ implements the ``Syzygy of Thoughts'' mechanism from Section~\ref{sec:method}. Each call to \textsc{LLMResolve}$(M, \Psi)$ yields a candidate reasoning chain $\mathcal{S}_k$ that satisfies the auxiliary conditions. From an algebraic viewpoint, each $\mathcal{S}_k$ functions as a syzygy relating different generators of $M$, and the list $\mathcal{M}$ collects all such candidates.
Selecting a minimal resolution (Minimality).  
Once the candidate set $\mathcal{M}$ is formed, the subroutine \textsc{ArgMinScore}$(\mathcal{M}, \phi)$ selects the best resolution according to a scoring function~$\phi$. This function, implemented via an LLM scoring prompt, evaluates both structural regularity (e.g., consistency with freeness and mapping constraints) and task-specific correctness. This step aligns with the ``Minimality'' module in Figure~\ref{fig:main}.
Extracting the final answer (Exactness).  
Finally, \textsc{ExtractFinalAnswer}$(\mathcal{S}_{\text{opt}}, D)$ converts the selected reasoning chain into the final task-specific answer. Pattern matching techniques (e.g., regular expressions or JSON parsing) ensure the output is consistent with the resolved structure and adheres to benchmark formatting requirements, realizing the ``Exactness'' principle.

\twocolumn

\subsection{Conceptual Illustration of the SoT Pipeline}\label{sec:appendix_sot_illustration}

\begin{figure*}
    \centering
    \includegraphics[width=\linewidth]{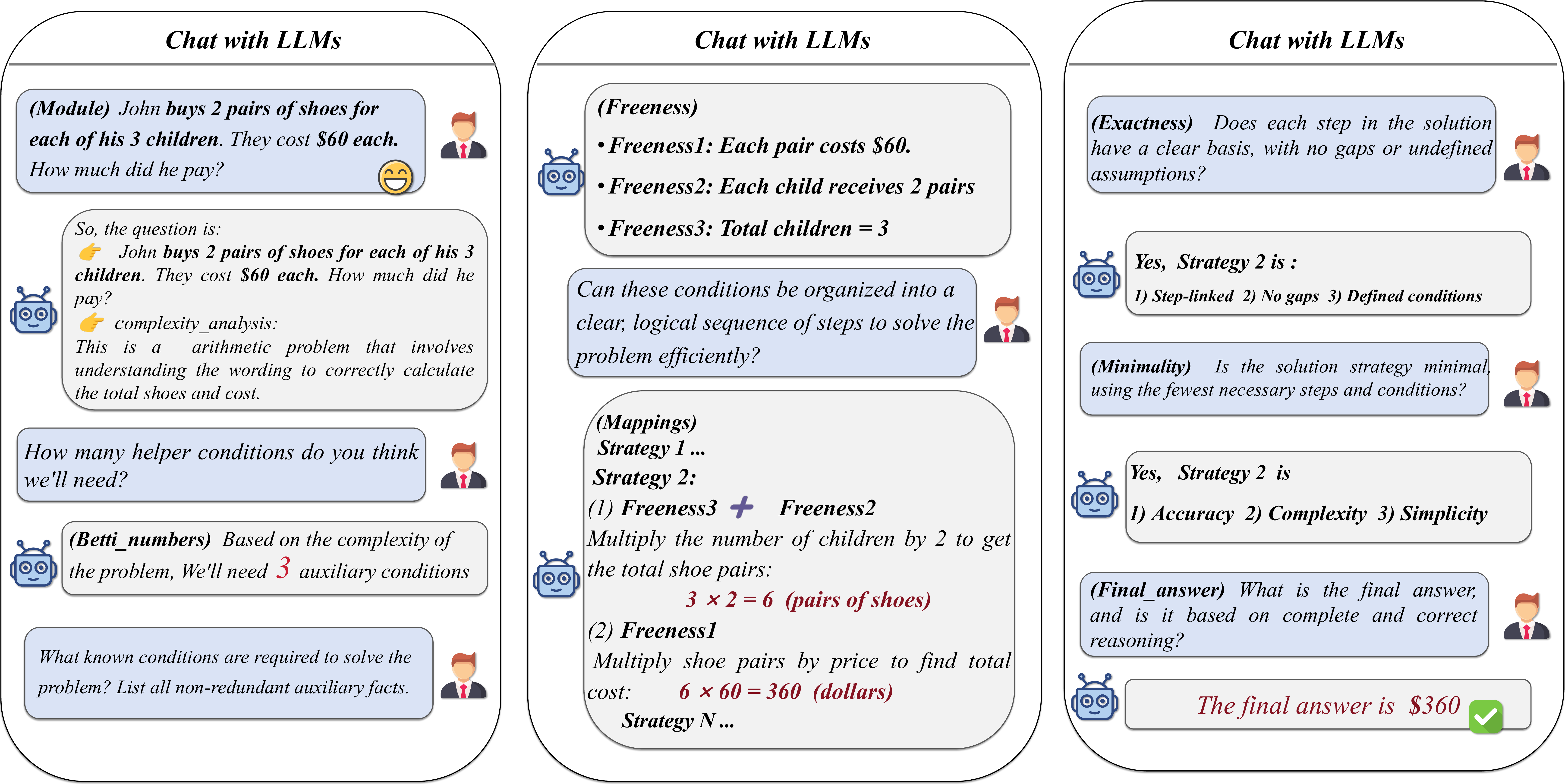}
    \caption{Conceptual framework illustrating the process of navigating LLMs' latent space through modular reasoning. The diagram highlights key components, including Module Freeness, Betti Numbers, Mappings, Exactness, and Minimality, to decompose and solve a complex problem. While the figure aids in understanding the theoretical underpinnings and logical flow of the methodology, it does not represent the exact procedural steps of the method.}
    \label{fig:conceptual}
\end{figure*}

\paragraph{Explanation.}
Figure~\ref{fig:conceptual} provides a concrete walk-through of the SoT framework using a simple arithmetic problem. 
The illustration makes explicit how SoT transforms a natural-language question into a structured multi-stage reasoning process.

\noindent\textbf{(1) Module and Betti Numbers.}
The LLM first rewrites the question into a normalized “module” form and performs a lightweight complexity analysis. 
Based on the estimated problem structure, it predicts the required number of auxiliary conditions, corresponding to the Betti-number estimate~$\beta$ discussed in Section~\ref{sec:method}.
\textbf{(2) Freeness.}
The LLL generates non-redundant auxiliary conditions (e.g., price per pair, number of children, number of pairs per child). 
These auxiliary facts form the initial free modules $F_0, F_1, \dots$, which constrain the search space of valid reasoning chains.
\textbf{(3) Mappings.}
The LLM organizes auxiliary conditions into candidate reasoning strategies by combining freeness conditions in different orders. 
Each such combination corresponds to a candidate syzygy, i.e., a structured reasoning chain that reflects different possible mappings between the modules in the minimal free resolution.
\textbf{(4) Exactness.}
The LLM inspects each candidate strategy and verifies whether every reasoning step is justified, gap-free, and based on explicitly stated assumptions. 
This matches the exactness requirement in the algebraic SoT formulation.
\textbf{(5) Minimality.}
Among all valid strategies, the LLM selects the one with the smallest reasoning footprint, balancing accuracy, complexity, and simplicity. 
This corresponds to the minimality principle described in Section~\ref{sec:method}.
\textbf{(6) Final Answer.}
Once the optimal syzygy is chosen, the LLM extracts the final numerical answer while ensuring consistency with all earlier reasoning stages. 
This mirrors the final extraction step in Algorithm~\ref{alg:minimal_free_resolution_standard}.

\paragraph{Relation to Algorithm~\ref{alg:minimal_free_resolution_standard}.}
This example illustrates the same pipeline implemented formally in Algorithm~\ref{alg:minimal_free_resolution_standard}: 
\textsc{Analyse}$(M, D)$ corresponds to Step 1; 
\textsc{LLMGenerateAuxCondition} and \textsc{LLMResolve} correspond to Steps 2--3; 
\textsc{ArgMinScore} corresponds to Step 4; 
and \textsc{ExtractFinalAnswer} corresponds to Step 5.
The illustration provides an intuitive view complementary to the algorithmic description.

\subsection{Complex Reasoning Evaluation}\label{sec:complex}

To compare the complex problem reasoning capabilities of SoT with mainstream methods, we evaluated SoT and compared it with extensive open-source data from various papers~\cite{chen2025towards}. Specifically, we applied the SoT reasoning chain to GPT-4o-mini and tested its reasoning accuracy on the GSM8K and MATH datasets, while comparing it with other methods from the literature.\\
\textbf{Experimental Configuration:} The test model was GPT-4o-mini, with the Betti number set to 7 and mapping set to 3. The baselines included CoT and various popular chain-of-thought variants (such as MathPrompter, QuaSAR, MathDivide, etc.). The datasets used were GSM8K ~\cite{cobbe2021training} and MATH ~\cite{hendrycksmath2021}.

\begin{table}[h]
\centering
\scalebox{0.5}{
\begin{tabular}{l|c|c|c}
\toprule
 \textbf{Method} & \textbf{Model} & \textbf{GSM8k} & \textbf{MATH} \\ 
\midrule
No-CoT~\cite{deng2024explicit} & Mistral-7B   & 38.0\% & - \\
ICoT-SI~\cite{deng2024explicit} & Mistral-7B  & 51.0\% & - \\
- & RecurrentBlock-3.5B   & 42.1\% & - \\
MathCoder-CL~\cite{wang2023mathcoder} & Code-Llama-7B   & 67.8\% & 30.2\% \\
MAmmoTH~\cite{yue2023mammoth} & Code-Llama-7B   & 59.4\% & - \\
Brain~\cite{chen2024brain} & Code-Llama-7B   & 74.0\% & - \\
SQ-VAE~\cite{wang2023guiding} & Llama-2-7B   & 40.0\% & 7.0\% \\
Self-Rewarding~\cite{chen2024self} & Llama-2-7B   & 40.0\% & 10.7\% \\
STaR~\cite{zelikman2022star} & Llama-2-7B   & 58.2\% & 16.0\% \\
ENVISIONS~\cite{xu2024interactive} & Llama-2-7B   & 59.0\% & 19.0\% \\
MetaMath~\cite{yu2023metamath} & Llama-2-7B   & 66.5\% & - \\
ToRA-Code~\cite{gou2023tora} & Llama-2-7B   & 72.6\% & - \\
OVM~\cite{yu2023ovm} & Llama-2-7B  & 73.7\% & - \\
- & Llama-3.1-8B   & 56.7\% & 20.3\% \\
- & Llama-3.1-70B & 85.5\% & 41.4\% \\
- & Llama-3.1-405B & 89.0\% & 53.8\% \\
- & NuminaMath-7B-CoT & 75.4\% & 55.2\% \\
- & DeepSeek-Coder-7B & 77.4\% & 44.4\% \\
- & Qwen2-7B & 79.9\% & 44.2\% \\
- & Qwen2-Math-7B & 80.4\% & 50.4\% \\
SIaM~\cite{yu2024siam} & Qwen-2-Math-Base   & 81.5\% & 50.0\% \\
- & Internlm2-math-plus-7B & 84.0\% & 54.4\% \\
OMI2~\cite{li2025codei} & Qwen2.5-Coder-7B  & 84.1\% & 72.3\% \\
CODEI/O++~\cite{li2025codei} & Qwen2.5-Coder-7B   & 85.7\% & 72.1\% \\
PyEdu~\cite{li2025codei} & Qwen2.5-Coder-7B   & 85.8\% & 71.4\% \\
CODEI/O~\cite{li2025codei} & Qwen2.5-Coder-7B   & 86.4\% & 71.9\% \\
OC-SFT-1~\cite{li2025codei} & Qwen2.5-Coder-7B   & 86.7\% & 70.9\% \\
WI~\cite{li2025codei} & Qwen2.5-Coder-7B   & 87.0\% & 71.4\% \\
WI (Full)~\cite{li2025codei} & Qwen2.5-Coder-7B   & 87.0\% & 71.1\% \\
OMI2 (Full)~\cite{li2025codei} & Qwen2.5-Coder-7B    & 88.5\% & 73.2\% \\
- & DeepSeekMath-7B-RL & 88.2\% & 51.7\% \\
\midrule
CoMAT~\cite{leang2024comat} & GPT-4 & 93.7\% & - \\
CoT~\cite{ranaldi2025improving} & GPT-4 & 94.5\% & - \\      
FCoT~\cite{lyu2023faithful} & GPT-4 & 95.0\% & - \\
MathPrompter~\cite{imani2023mathprompter} & GPT-4 & 95.6\% & - \\
QuaSAR~\cite{radford2018improving} & GPT-4 & 96.5\% & - \\
MathDivide~\cite{srivastava2024mathdivide} & GPT-4 & 96.8\% & - \\
        \textbf{SoT (Ours)} & \textbf{GPT-4o-mini}  & \textbf{96.0\%} & \textbf{79.1\%} \\
\bottomrule
\end{tabular}
}
\caption{Performance on Different Benchmarks}
\label{tab:test01}
\end{table}

\noindent
\textbf{Experimental Results:} As shown in Table ~\ref{tab:test01}, SoT achieved 96.0\% on GSM8K and 79.1\% on MATH on GPT-4o-mini. On GSM8K, SoT's performance is close to the best result achieved by GPT-4 (e.g., MathDivide's 96.8\%) ~\cite{chen2025towards} and outperforms the best 7B model (e.g., OMI2 Full's 88.5\%)~\cite{chen2025towards}. On MATH, SoT’s 79.1\% significantly outperforms the best mainstream 7B models (e.g., OMI2 Full's 73.2\%). This result indicates that SoT can achieve reasoning capabilities close to those of closed-source large models on lightweight models, significantly narrowing the performance gap between open-source models and GPT-4, validating its superiority in complex mathematical reasoning.

\subsection{Path Counting and Token-Cost Metrics}
\label{sec:path_define}

To align the notion of ``paths'' across different methods, we adopt a unified definition. 
For a method $m$ and a sample $x$, let 
$N_{\mathrm{cand}}^{(m)}(x)$ denote the number of 
candidate reasoning trajectories generated on $x$ that contain a complete chain of reasoning and yield a directly scorable answer. 
The horizontal axis in our plots reports
\[
\mathrm{AvgPaths}(m) 
= \frac{1}{|D|}\sum_{x \in D} N_{\mathrm{cand}}^{(m)}(x),
\]
which reflects how many full candidate solutions, on average, the method explores per problem.

Under this formulation:

\begin{itemize}
    \item \textbf{CoT:} Each problem produces exactly one chain-of-thought, hence $\mathrm{AvgPaths} \approx 1$.

    \item \textbf{CoT-SC($n$):} Each problem samples $n$ independent CoT trajectories followed by majority voting, thus $\mathrm{AvgPaths} \approx n$.

    \item \textbf{GoT($b,L$):} During multi-round graph search, each invocation that produces a complete candidate solution is counted as a path. 
    Internal graph-expansion calls that generate intermediate nodes without a final answer do not count as paths, though their token usage is included in the bubble size.

    \item \textbf{AoT($d$):} Along the Markov chain, solving each of $\{Q_i, G_i, Q_{i+1}\}$ produces a directly scorable candidate answer. 
    Hence each \texttt{solve($\cdot$)} call is counted as one path, and $\mathrm{AvgPaths}$ approximates the average number of solver calls per Markov chain.

    \item \textbf{SoT($\beta,\mu$):} In SoT, \texttt{LLMRESOLVE} generates $\mu$ syzygies $\{S_k\}_{k=1}^{\mu}$, each constituting a full structured reasoning chain capable of producing an answer. 
    Therefore the number of candidate paths is exactly the number of syzygies:
    \[
    N_{\mathrm{cand}}^{(\mathrm{SoT})}(x) = \mu.
    \]
    The $\beta$ auxiliary calls used during the freeness stage do not count as paths, but all of their token usage is included in the overall cost.
\end{itemize}

The bubble area in the figure corresponds to the average tokens per problem, computed as the total number of tokens consumed by all LLM calls (generation, intermediate analysis, candidate solutions, and scoring), averaged over the evaluation set. This metric reflects the overall reasoning cost of each method.

\end{document}